\documentclass[acmlarge]{acmart}
\AtBeginDocument{%
  }

\setcopyright{acmlicensed}
\copyrightyear{2018}
\acmYear{2018}
\acmDOI{XXXXXXX.XXXXXXX}

\acmJournal{POMACS}
\acmVolume{37}
\acmNumber{4}
\acmArticle{111}
\acmMonth{8}

\usepackage{amsmath}
\usepackage{epsfig}

\usepackage{tabu}                      % only used for the table example
\usepackage{booktabs}                  % only used for the table example
\usepackage{lipsum}                    % used to generate placeholder text
\usepackage{mwe}                       % used to generate placeholder figures

\usepackage[utf8]{inputenc} % allow utf-8 input
\usepackage[T1]{fontenc}    % use 8-bit T1 fonts
\usepackage{multirow}
\usepackage{subfig}
\usepackage{amsfonts}       % blackboard math symbols
\usepackage{nicefrac}       % compact symbols for 1/2, etc.
\usepackage{microtype}      % microtypography
\usepackage{graphicx}
\usepackage{subcaption} % 需要加载 subcaption 包
\usepackage{tabularx}
\usepackage{xcolor} 
\usepackage[table]{xcolor}  % 允许表格着色
\usepackage{algorithm}
\usepackage{algpseudocode}
\usepackage{tikz}
\usepackage{hyperref}
\usepackage{listings}
\newfloat{Code}{t}{lop}
\floatname{Code}{Code Snippet}
\restylefloat{Code}
\definecolor{red}{rgb}{0.75, 0.0, 0.0}

\begin{document}

\title{Data-Efficient Brushstroke Generation with Diffusion Models for Oil Painting}

\author{Dantong Qin}
\email{D.Q.Dantong@tudelft.nl}
\orcid{0009-0000-7746-0863}
\affiliation{%
  \institution{Delft University of Technology}
  \city{Delft}
  \country{Netherlands}}

\author{Alessandro Bozzon}
\email{A.Bozzon@tudelft.nl}
\affiliation{%
  \institution{Delft University of Technology}
  \city{Delft}
  \country{Netherlands}
}

\author{Xian Yang}
\email{xian.yang@manchester.ac.uk}
\affiliation{%
  \institution{The University of Manchester}
  \city{Manchester}
  \country{UK}
}

\author{Xun Zhang}
\email{x.zhang-16@tudelft.nl}
\affiliation{%
 \institution{Delft University of Technology}
 \city{Delft}
 \country{Netherlands}
}

\author{Yike Guo}
\email{yikeguo@ust.hk}
\affiliation{%
  \institution{The Hong Kong University of Science and Technology}
  \city{Kowloon}
  % \state{Beijing Shi}
  \country{HKSAR}}

\author{Pan Wang}
\email{P.Wang-2@tudelft.nl}
\authornote{Corresponding author to this work.}
\affiliation{%
  \institution{Delft University of Technology}
  \city{Delft}
  % \state{Texas}
  \country{Netherlands}}

\renewcommand{\shortauthors}{Qin et al.}

\begin{abstract}
\label{sec:abstract}
Many creative multimedia systems are built upon visual primitives such as strokes or textures, which are difficult to collect at scale and fundamentally different from natural image data. This data scarcity makes it challenging for modern generative models to learn expressive and controllable primitives, limiting their use in process-aware content creation. In this work, we study the problem of learning human-like brushstroke generation from a small set of hand-drawn samples ($n=470$) and propose StrokeDiff, a diffusion-based framework with Smooth Regularization (SmR). SmR injects stochastic visual priors during training, providing a simple mechanism to stabilize diffusion models under sparse supervision without altering the inference process. We further show how the learned primitives can be made controllable through a Bézier-based conditioning module and integrated into a complete stroke-based painting pipeline, including prediction, generation, ordering, and compositing. This demonstrates how data-efficient primitive modeling can support expressive and structured multimedia content creation. Experiments indicate that the proposed approach produces diverse and structurally coherent brushstrokes and enables paintings with richer texture and layering, validated by both automatic metrics and human evaluation.
\end{abstract}

\begin{CCSXML}
<ccs2012>
   <concept>
       <concept_id>10010405.10010469.10010470</concept_id>
       <concept_desc>Applied computing~Fine arts</concept_desc>
       <concept_significance>500</concept_significance>
       </concept>
   <concept>
       <concept_id>10010147.10010178.10010224</concept_id>
       <concept_desc>Computing methodologies~Computer vision</concept_desc>
       <concept_significance>500</concept_significance>
       </concept>
 </ccs2012>
\end{CCSXML}

\ccsdesc[500]{Applied computing~Fine arts}
\ccsdesc[500]{Computing methodologies~Computer vision}

\keywords{Brushstroke Synthesis, Stroke-based Rendering, Diffusion Models, Artist Painting, Few-shot}

\received{20 February 2007}
\received[revised]{12 March 2009}
\received[accepted]{5 June 2009}

\maketitle

\begin{figure*}[t]
    \centering
    \includegraphics[width=1\linewidth]{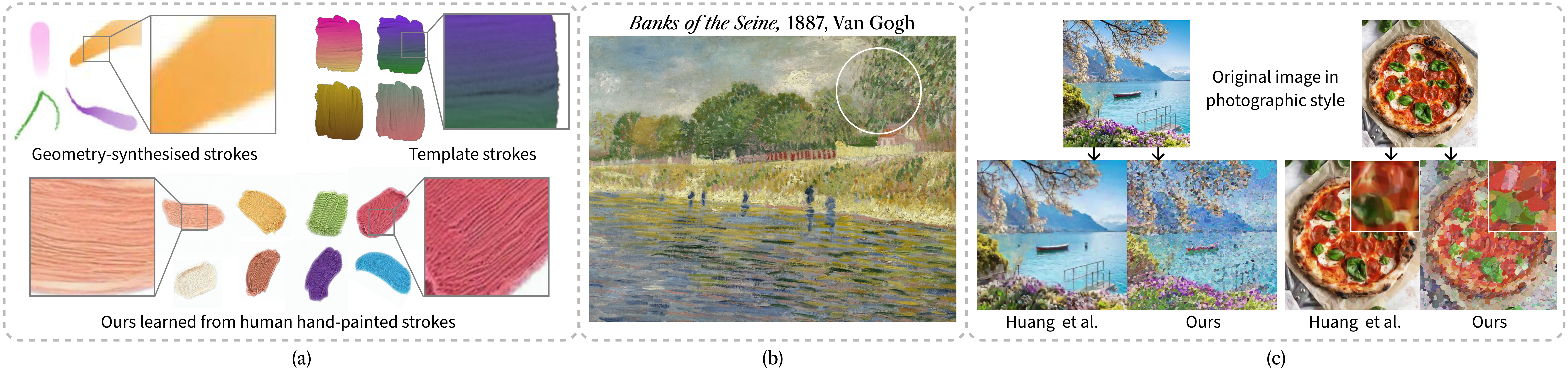}
    \caption{Comparison of stroke quality and painting outputs.  
(a) Prior SBR methods use geometric or template strokes with limited expressive capacity. Our strokes, learned from human-painted data, yield richer structure and texture. (b) Target oil painting style, showing abstraction beyond realism. 
(c) Compared to Huang et al. \cite{huang2019learning}, 
% which fits geometric shapes to reproduce the original photographic content, the approach applied with our strokes introduces a more pronounced artistic domain shift.
Our method introduces a stronger artistic domain shift.} 
    \label{fig:first}
\end{figure*}

\section{Introduction}
\label{sec1}

Painting is a layered, iterative process in which artists build visual scenes through successive brushstrokes, each carrying spatial, textural, and stylistic intent \cite{cetinic2022understanding,li2025immersive,wang2025ai}. In creative multimedia systems, stroke-based rendering (SBR) is a representative paradigm that seeks to emulate this process by synthesizing images as sequences of discrete brushstrokes. Unlike photorealistic synthesis, SBR does not aim to reproduce pixels but to convey semantic content through painterly representation. Recent methods have made progress in generating content-aligned renderings, yet most rely on predefined stroke templates \cite{hu2024towards,tang2024attentionpainter,liu2021paint, tong2022im2oil, schaldenbrand2021content} or geometry-synthesized primitives \cite{huang2019learning, singh2021combining,nakano2019neural, wang2023stroke}. While these approaches offer efficiency and structural consistency, they often struggle to capture the diversity, irregularity, and texture of human brushstrokes, which are essential for stylistic abstraction. As a result, the expressive space of human-like strokes remains underexplored in data-driven painting pipelines (see Fig. \ref{fig:first} for comparison).

To address this gap, we investigate the problem of generating human-like brushstrokes using diffusion models (DMs). Compared to natural image domains, where large-scale datasets are widely available, real brushstroke data remain scarce and underexplored in existing literature. Compared to earlier approaches such as GAN-based stroke synthesis \cite{nakano2019neural, wang2023stroke,kotovenko2021rethinking}, DMs excel when trained on abundant data, but their ability to generalize to design-time primitives is less understood, and direct fine-tuning of pretrained models on a few hundred samples tends to result in structural degradation. This highlights the challenge of stabilizing diffusion training under limited data while maintaining the abstraction and stylistic qualities of artistic strokes.

Previous efforts to address low-resource diffusion have either modified the diffusion path (e.g., noise scheduling \cite{gu2022f, hoogeboom2023simple, chen2023importance,lin2024common, NEURIPS2024_7eb6233e, NEURIPS2024_40eff167}, classifier (or classifier-free) guidance \cite{dhariwal2021diffusion,ho2022classifier, NEURIPS2023_b87bdcf9}) or inject external conditions such as prompts \cite{dong2023prompt,nam2024optical} or exemplars\cite{kumari2023multi,ruiz2023dreambooth,yan2025neural,hu2024instruct}. While effective in some domains, these strategies often require large-scale tuning or introduce test-time dependencies, which limits their scalability. In this work, we propose \textbf{StrokeDiff}, a diffusion-based framework designed for learning expressive visual primitives under extremely limited data. At its core is \textbf{Smooth Regularization (SmR)}, a training-time strategy that injects \emph{stochastic visual priors} into the forward diffusion process to compensate for the lack of structural supervision. By injecting weak but diverse visual cues during training, SmR encourages the model to preserve global structure and semantic coherence in low-signal regimes. Importantly, SmR operates purely at training time and does not modify the inference procedure or require additional conditions. This makes it a lightweight and general regularizer for stabilizing diffusion models under sparse data, without introducing test-time dependencies or computational overhead.

Beyond individual strokes, to demonstrate that the learned primitives can support structured content creation, we further introduce a Bézier-parametrized conditioning module that enables controllable stroke synthesis and facilitates integration into painting pipelines. This design provides explicit control over stroke attributes such as shape and placement, and supports their arrangement into complete images. To further improve composition, we add a ranking loss that regularizes stroke order, mitigating overlap artifacts and enhancing stylistic layering.

Our experiments indicate that StrokeDiff generates high-fidelity and diverse strokes from 470 hand-drawn samples, generalizes better than noise scheduling and LoRA-based adaptation under sparse data. At the level of complete paintings, it produces more coherent and stylistically expressive results, which we evaluate using both semantic metrics such as CLIP alignment and human preference studies that assess style, texture, and overall aesthetic quality. Our main contributions are:
\begin{itemize}
    \item Smooth Regularization (SmR): a training-time regularization strategy based on stochastic visual prior injection, which stabilizes diffusion training under sparse data without modifying the inference procedure.
    \item Controllable stroke synthesis and pipeline integration: we design a Bézier-parametrized conditioning module for controllable stroke generation, and connect it with a complete painting pipeline, leading to brushstroke renderings that better capture the layering and textural qualities of oil painting.
    \item Evaluation: we conduct multi-dimensional evaluation at both the stroke and painting levels, using CRD, FID, LPIPS, and human studies to compare template-, vector-, GAN-, and diffusion-based approaches.
\end{itemize}

\section{Related work}
\label{sec:relatedwork}
\subsection{Strokes in Stroke-Based Rendering}
\label{sec:relatedwork1}

While generative models have also made progress in image stylization~\cite{huang2025creativesynth,niu2024multi,wang2022multi}, stroke-based rendering constructs images through sequential brushstrokes, emulating the layered approach of human artists \cite{huang2019learning, singh2022intelli, singh2021combining,liu2023painterly,ibarrola2023collaborative}. Existing methods can be broadly categorized into template-based, neural stroke generation, and vector graphics-based approaches.
Template-based approaches \cite{tong2021sketch, liu2021paint, tong2022im2oil, schaldenbrand2021content, hu2023stroke,hu2024towards} transform a limited set of stroke templates through parameter adjustments (e.g., color, length, and angle). 
% Liu et al. \cite{liu2021paint} employed a DETR-based \cite{carion2020end} network to predict stroke parameters from reference images, which were then rendered with fixed templates. Although this strategy can effectively reproduce image content, it often yields uniform strokes lacking the rich textures of real brushwork.
While they preserve the appearance of painting textures, their reliance on fixed templates often leads to repetitive stroke patterns.
Neural stroke generation methods \cite{nakano2019neural, wang2023stroke, zou2021stylized} aim to increase stroke diversity by synthesizing strokes from noise. However, many of these models are trained on synthetic datasets (e.g., MyPaint\footnote{\url{https://github.com/mypaint/mypaint}} or FluidPaint\footnote{\url{https://david.li/paint/}}), leading to overly smooth and geometric outputs. Although Bidgoli et al.~\cite{bidgoli2020artistic} trained a VAE on real brushstrokes, their method was not integrated into a complete painting pipeline. Another line of work employs vector graphics-based approaches \cite{li2020differentiable,huang2019learning,hu2024vectorpainter, chen2025spline} to fit brushstrokes as parametric curves, such as Bézier splines. While these methods achieve strong structural fidelity, they often sacrifice the spontaneity and organic variation found in freehand painting. Despite advances in SBR, achieving diverse, realistic, and expressive brushstrokes remains a core challenge.

\subsection{Low-data diffusion methods}

Recent work has adapted diffusion models to low-data regimes by incorporating external priors or prompts to guide generation, including exemplar-based personalization methods such as DreamBooth~\cite{ruiz2023dreambooth} and Custom Diffusion~\cite{kumari2023multi}, or inversion-based approach~\cite{mokady2023null}, and attention manipulation techniques~\cite{hertz2022prompt,sohn2023styledrop}. While effective for high-fidelity or controllable generation, these methods typically rely on test-time inputs or prompt engineering, and are often constrained to instance-specific synthesis. Another line of research focuses on improving sample diversity under constrained training by modifying the diffusion path through noise scheduling~\cite{gu2022f, hoogeboom2023simple, chen2023importance, lin2024common}, or posterior guidance~\cite{ho2022classifier}, or multi-stage refinement frameworks such as DoD~\cite{yue2024diffusion}, which incorporate visual priors from earlier stages to enhance semantic structure and fidelity. Our method introduces a training-time modification to the diffusion path yet maintains no test-time conditions.

\section{Method}

\begin{figure*}[t]
    \centering
    \includegraphics[width=1\linewidth]{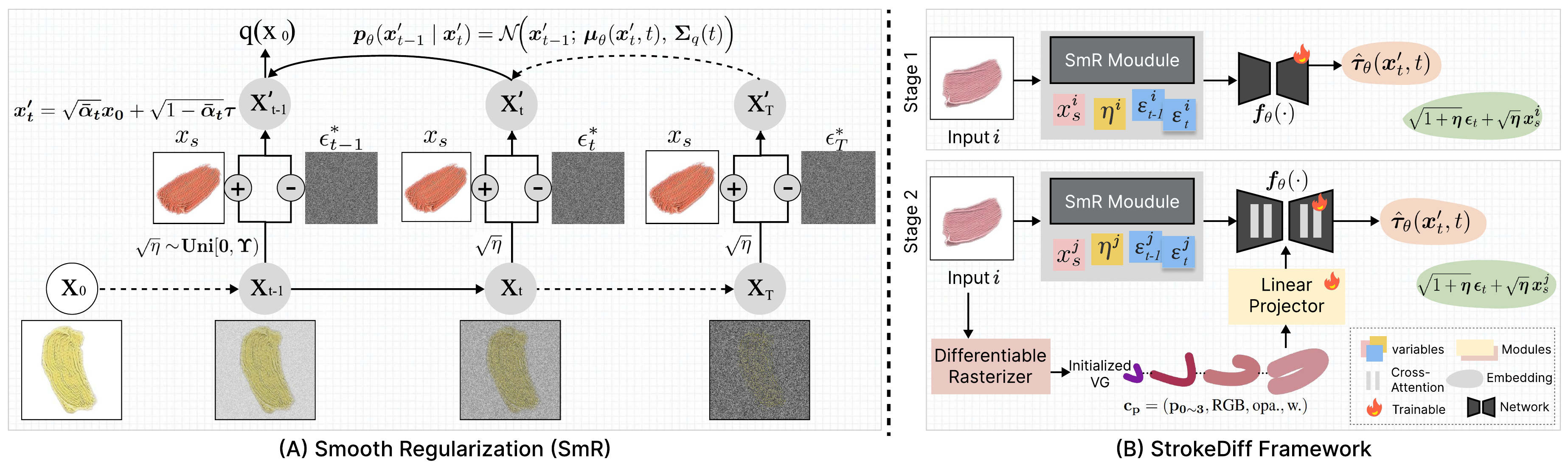}
    \caption{Overview of our method. (A) Smooth Regularization injects priors during training to mitigate mode collapse under limited data. At each timestep, a brushstroke $x_s$ is sampled from the dataset and injected into the forward process alongside a newly sampled noise term $\epsilon^* \sim \mathcal{N}(0, I)$; both terms are scaled by a factor $\eta \sim \mathrm{Uni}[0, \Upsilon]$. (B) The StrokeDiff framework is trained in two stages. In Stage 1, the model learns to generate strokes with SmR but without conditional inputs. In Stage 2, we incorporate parameter conditioning via a raster-to-vector module that predicts vectorized stroke parameters (e.g., control points, opacity, RGB, width). These parameters are then used to control the generation of specific brushstroke attributes.}
    \label{fig:model}
    \vspace{-1mm}
\end{figure*}

Our goal is to train a diffusion model on a small set of hand-drawn brushstrokes, enabling it to generate strokes that are human-like in both texture and structure. The proposed framework, \textit{StrokeDiff}, augments standard latent diffusion with visual prior injection and stroke-level controllability (see Fig. \ref{fig:model} for an overview). We begin by training the model without inference-time conditioning, and introduce \textit{Smooth Regularization (SmR)} to inject weak visual priors during training, mitigating the mode collapse observed when fine-tuning a latent diffusion model (LDM) by modifying the diffusion process. SmR is applied only during training and deactivated at inference, which allows standard noise sampling without additional preparation. To make the outputs compatible with downstream tasks, we further parameterize brushstrokes and support controllable generation based on stroke attributes. These components together allow StrokeDiff to serve both as a stand-alone stroke generator and as the foundation of a broader painting pipeline, which we detail later in this section.
 
% SmR is designed to be removed at inference time by reverting to standard noise sampling, avoiding the need for external or predefined conditions that are impractical for painting tasks. 

\subsection{Preliminaries}
Diffusion models generate samples by iteratively denoising an initial Gaussian noise input. In the forward process, a clean image \(x_0\) is gradually corrupted by adding Gaussian noise in a Markov chain. One way to express this is:
\begin{equation}
\small
    q(x_t \mid x_{t-1}) 
    = \mathcal{N}\!\Bigl(x_t;\,\sqrt{\alpha_t}\,x_{t-1},\,(1-\alpha_t)I\Bigr),
\end{equation}
where \(\alpha_t\) controls how much information from \(x_{t-1}\) is retained. By applying this step repeatedly from \(t=1\) to \(T\), we can write a closed-form distribution conditioned on \(x_0\):
\begin{equation}
\small
    q(x_t \mid x_0) 
    = \mathcal{N}\!\Bigl(x_t;\,\sqrt{\bar{\alpha}_t}\,x_0,\,(1-\bar{\alpha}_t)I\Bigr),
\end{equation}
where \(\bar{\alpha}_t = \prod_{i=1}^t \alpha_i\). 

To invert this process, a neural network \(\epsilon_\theta(x_t,t)\) is trained to predict the noise \(\epsilon \sim \mathcal{N}(0,I)\) added to \(x_0\), minimizing
\begin{equation}
\small
    \mathcal{L}(\theta) 
    = \mathbb{E}_{x_0,\epsilon,t}\!\bigl[\|\epsilon - \epsilon_\theta(x_t,t)\|^2\bigr].
\end{equation}
Using these predictions, the reverse transition is:
\begin{equation}
\small
    p_\theta(x_{t-1} \mid x_t) 
    = \mathcal{N}\!\Bigl(x_{t-1};\,\mu_\theta(x_t,t),\,\Sigma_q(t)\Bigr),
\end{equation}
where \(\mu_\theta(x_t,t)\) and \(\Sigma_q(t)\) guide the reconstruction of \(x_0\).

\begin{figure}[t]
    \centering
    \includegraphics[width=1\linewidth]{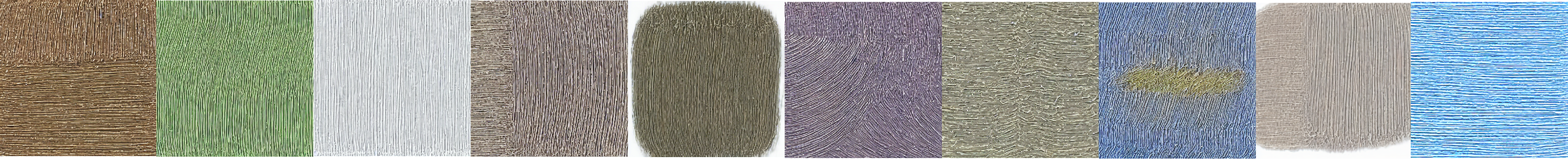}
    \caption{Stroke generation using Stable Diffusion v1-5 \cite{rombach2022high} fine-tuned on typical stroke data that lacks sufficient diversity. Due to severe mode collapse early in training, the model produces only texture-like images with minimal variation and no authentic brushstroke resemblance.}
    \label{fig:navie}
    \vspace{-3mm}  % 让正文更靠近图片
\end{figure}

\subsection{High-fidelity stroke generation with visual guidance}

\subsubsection{Problem Setting}Our work focuses on learning to generate brushstroke primitives collected from a single artist, comprising 470 scanned strokes. These primitives represent process-oriented visual elements that are rarely observed independently in public datasets, which are fundamentally out-of-distribution for open-domain latent diffusion models, making them unrecoverable via prompt tuning or conditional guidance. The direct fine-tuning of Stable Diffusion 1.5 on this dataset leads to severe mode collapse (see Fig. \ref{fig:navie}). While the strokes exhibit family-level stylistic coherence, they are semantically sparse and structurally diverse, leading to a low information density problem. As a result, the fine-tuned model fails to infer global structure during the early denoising steps. Given the low-frequency structures could require longer temporal horizons to learn \cite{hoogeboom2023simple}, we hypothesize that the weak semantic signal is rapidly overwhelmed by noise under standard diffusion schedules. To address this, we propose a regularization strategy to facilitate structure preservation in low-signal regimes.

\subsubsection{Stochastic Visual Prior}
We introduce a \emph{visual prior} \(x_s\) at each diffusion step. Rather than relying solely on noisy reconstructions of the target stroke, we sample \(x_s\) a random brushstroke from the training set, leveraging inter-stroke consistency to provide structure-aware guidance. The modified forward process is defined as:
\begin{equation}
\small
x_t' \;=\;
\fcolorbox{red}{white}{$x_t$}
\;+\;\fcolorbox{yellow}{white}{$\sqrt{1-\bar{\alpha}_t}\,\sqrt{\eta}\,x_s$}
\;-\;\fcolorbox{green}{white}{$\sqrt{1-\bar{\alpha}_t}\,\sqrt{\eta}\,\epsilon_t^*$},
\end{equation}
where \(\eta\) is a scaling factor (sampled from \(\mathrm{Uni}[0,\Upsilon)\), \(0<\Upsilon<1\)) that modulates the strength of prior injection. It is sampled once per training instance and shared across timesteps. Notably, 
\textsuperscript{\tikz \fill[red] (0,0) circle (2pt);}\(x_t'\) does not follow from \(x_{t-1}'\), 
\textsuperscript{\tikz \fill[yellow] (0,0) circle (2pt);}\(x_t\) is interpolated with \(x_s\) at each timestep, and 
\textsuperscript{\tikz \fill[green] (0,0) circle (2pt);}\(\epsilon^*\sim \mathcal{N}(0,I)\) is a newly sampled noise added to maintain diversity by increasing overall variance. Otherwise, the additional structure would reduce the noise magnitude and potentially bias training. The resulting distribution $q(x_t' \mid x_0) $ includes both a data term \(\sqrt{\bar{\alpha}_t}\,x_0\) and a visual prior term \(\sqrt{1-\bar{\alpha}_t}\,\sqrt{\eta}\,x_s\), as shown below:
\begin{align}
= \mathcal{N}\Bigl(
x_t';\,
\underbrace{\sqrt{\bar{\alpha}_t}\,x_0}_{\text{Data Term}}
+
\underbrace{\sqrt{1-\bar{\alpha}_t}\,\sqrt{\eta}\,x_s}_{\text{Visual Prior}},
\underbrace{(1+\eta)\,(1-\bar{\alpha}_t)\,\mathbf{I}}_{\text{Variance Term}}
\Bigr).
\end{align}
\normalsize

\noindent The visual prior \(x_s\) does not propagate across timesteps. As a result, the injected structure remains temporally local and does not accumulate through the trajectory. This non-propagating nature prevents the model from memorizing fixed structural patterns, in contrast to deterministic scheduling schemes where the noise trajectory is predefined across steps and samples. 

% , and this non-deterministic injection avoids accumulating prior information, preventing overfitting to a fixed noise pattern. It also introduces new structural cues at each step, boosting stroke diversity while preserving essential brushstroke semantics.

We derive a modified posterior distribution that $q(x_{t-1}' | x_t', x_0)$
\small
\begin{align}
   \propto \mathcal{N} \Bigg( x_{t-1}';  
    &\frac{\sqrt{\alpha_t}(1-\bar{\alpha}_{t-1})(1+\eta)x_t' + \sqrt{\bar{\alpha}_{t-1}} ((1-\alpha_t-2\bar{\alpha}_t)\eta + 1-\alpha_t )x_0 + f(\eta)}
    {1-\bar{\alpha}_t+(1+2\alpha_t -3\bar{\alpha}_t)\eta} ,  
    \nonumber \\ 
    &\hspace{-2em} \frac{\left((1+\alpha_t-2\bar{\alpha}_t) \eta + 1-\alpha_t\right)(1+\eta)(1-\bar{\alpha}_{t-1})}
    {1-\bar{\alpha}_t+(1+2\alpha_t -3\bar{\alpha}_t)\eta} \mathbf{I} \Bigg).
\end{align}
\normalsize
$f(\eta)$ represents the independent term dependent on $\eta$, expressed as:
\small
\begin{equation}
    f(\eta) = \sum_{k=1}^3 A_k \sqrt{\eta}^k,
\end{equation}
\normalsize
with coefficients $A_k$ determined by $\alpha_t$, $\alpha_{t-1}$, $\bar{\alpha}_t$, $\bar{\alpha}_{t-1}$, and $x_s$. 

\subsubsection{Smooth Regularization}With this formulation, we define $\tau$ as an alternative representation of the noise process:
\begin{equation}
    \tau = \sqrt{1+\eta}\epsilon + \sqrt{\eta}x_s,
\end{equation}
% \normalsize
% \begin{equation}
%     \boldsymbol{\tau} = \sqrt{1 + \boldsymbol{\eta}}\, \boldsymbol{\epsilon}_t + \sqrt{\boldsymbol{\eta}}\, \boldsymbol{x}_s
% \end{equation}
which encapsulates both learned noise and the injected prior. This allows us to parameterize the true denoising transition mean $\mu_q(x_t', x_0)$ as follows:
\footnotesize
\begin{align}
    \frac{
    \sqrt{\bar{\alpha}_{t-1}}(1 - \bar{\alpha}_t)x'_t 
    + C_1x'_t 
    - \sqrt{\bar{\alpha}_{t-1}}\sqrt{1 - \bar{\alpha}_t}(1 - \alpha_t)\tau 
    + C_2\tau + f(\eta)
}{
    \sqrt{\bar{\alpha}_t}\left(1-\bar{\alpha}_t+(1+2\alpha_t -3\bar{\alpha}_t)\eta \right)
},
\end{align}
% \normalsize
where $C_1$ and $C_2$ are coefficients, with each term multiplied with $\eta$. Since $\tau$ introduces additional stochasticity, the model must learn \textbf{both the underlying noise and the prior’s effect}. Thus, we parameterize $\hat{\tau}_\theta(x_t',t)$ via LDM's neural network, which refines the reconstruction process. Then, the optimization problem simplifies to:
\small
\begin{equation}
    \mathbf{L}_\tau(\theta) = \mathbb{E}\left[||\tau - \hat{\tau}_\theta(x_t',t)||^2\right].
\end{equation}
\normalsize
During inference, setting $\eta=0$ nullifies the visual prior, reducing to:
\begin{align}
    \mu_q(x_t', t) &= \mu_q(x_t, t) \text{\quad in DDPM},  &\\
    \Sigma_q(t) &= \Sigma_q(t) \text{\quad in DDPM}. 
\end{align}
This design remains compatible with existing diffusion sampling strategies, allowing the model to generate images purely from Gaussian noise without any additional priors. Since this approach “smooths” the noise via a modified training objective, we refer to it as \emph{Smooth Regularization}. In practice, it is straightforward to implement. Moreover, because \(x_s\) is randomly sampled, each \(x_0\) can pair with multiple distinct priors, effectively enlarging the training set and further boosting stroke diversity.

\subsection{Controllable stroke synthesis for downstream applications}
\label{sec:control}

Since our brushstrokes originate from scanned images, they lack an inherent parametric form. To introduce parametric control, we adopt a differentiable rasterizer \cite{li2020differentiable} that initializes a simple parametric primitive and iteratively refines it to match the target stroke. Specifically, we set both the path and segment count to 1, approximating each stroke with a single cubic B\'ezier curve:
\begin{equation}
    c_{p} \;=\; (p_0,\, p_1,\, p_2,\, p_3,\; \mathrm{R},\, \mathrm{G},\, \mathrm{B},\; \mathrm{opacity},\; \mathrm{width}),
\end{equation}
where each \(p_i\) holds the \((x,y)\) coordinates of a control point, resulting in \(c_p \in \mathbb{R}^{1\times13}\). By mapping strokes to this vector-graphic representation, we avoid manually recording brush trajectories \cite{bidgoli2020artistic}, thus making our approach compatible with any rasterized brushstroke data.

We integrate this parametric information into the model by concatenating \(c_p\) with a contextual feature \(c\) (e.g., color or style embedding) and project the combined vector:
\begin{equation}
    z \;=\; W\,[\,c_p;\,c\,],
\end{equation}
where \([\,\cdot;\,\cdot]\) denotes vector concatenation, and \(W\) is a learnable projection matrix. We then feed \(z\) into the U-Net’s cross-attention mechanism, fine-tuning both the projector and cross-attention layers. This setup enables conditional stroke generation that follows B\'ezier-defined shapes while preserving high-fidelity rasterization.

\begin{figure}[t]
    \centering
    % 第一个minipage：占0.48单栏宽度
    \begin{minipage}[t]{0.58\linewidth}
        \centering
        \includegraphics[width=\linewidth]{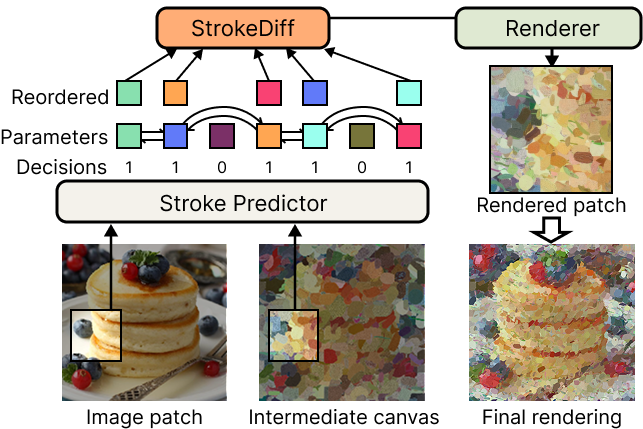}
        \caption{Illustration of the painting pipeline. Stroke parameters are reordered according to the predicted ranking scores $src_r$ to ensure a coherent rendering sequence.}
        \label{fig:pipeline}
    \end{minipage}
    \hfill  % 子图间空白
    % 第二个minipage：占0.48单栏宽度
    \begin{minipage}[t]{0.36\linewidth}
        \centering
        \includegraphics[width=\linewidth]{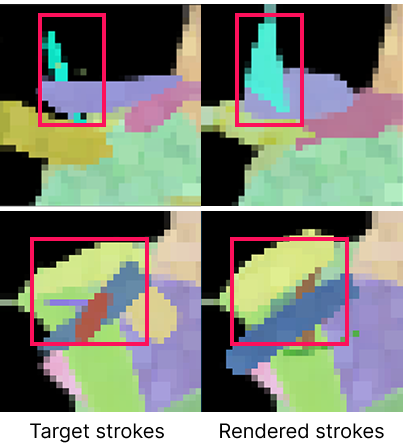}
        \caption{Strokes that are only optimally matched render an unordered sequence, leading to incorrect overlapping.}
        \label{fig:rank_stroke}
    \end{minipage}
\end{figure}

\subsection{Painting pipeline integration}
\label{sec:pipeline}

To render full paintings, we integrate StrokeDiff into a stroke-based rendering pipeline. The process involves (i) predicting stroke parameters (position, shape, size) from a target image, (ii) generating corresponding strokes using our diffusion model, and (iii) compositing them sequentially on a canvas with a renderer, as illustrated in Fig.~\ref{fig:pipeline}.

\noindent\textbf{Stroke predictor}. We adopt a DETR-style architecture similar to PaintTransformer \cite{liu2021paint}, which outputs a set of candidate strokes for each image region. Each stroke prediction is represented as \(\bigl(c_p,\; x_{\text{shift}},\; y_{\text{shift}},\; \text{scr}_r\bigr)\), where \(x_{\text{shift}},\,y_{\text{shift}}\) specify the placement offset, and \(\text{scr}_r\in(0,1)\) is a ranking score indicating the relative drawing order. For training, we first redefine a matching loss that aligns each predicted stroke with a ground-truth stroke under optimal assignment:
\begin{align}
\small
\mathcal{L}_{match} = \min_{\sigma} \sum_{i} \Big(
& \lambda_{\text{L1}} \,\|P^-_{i} - \hat{P}^-_{\sigma(i)}\|_1 + \lambda_{\text{cos}} \,\text{Cosine}(P^-_{i}, \hat{P}^-_{\sigma(i)}) + \lambda_{\text{D}} \,\text{BCE}(d_{i}, \hat{d}_{\sigma(i)}) \Big).
\end{align}
where \(P^-=\bigl(c_p,\; x_{\text{shift}},\; y_{\text{shift}})\) excludes the ranking score, $\sigma$ is the optimal bipartite matching, and $d \in \{0,1\}$ indicates stroke presence.

However, set-based prediction does not impose any ordering, which often leads to incoherent overlaps when strokes are drawn directly, as shown in Fig~\ref{fig:rank_stroke}. To address this, we introduce a ranking loss. Suppose the ground truth provides a sequence of strokes indexed as $\{1,2,…,vs\}$ in their correct drawing order, where a smaller index means the stroke should be painted earlier. For each valid sequence, the predictor outputs $\{\,\text{scr}_{r,u}\}_{u=1}^{vs}$, we regularize the predicted scores:
\begin{equation}
    \mathcal{L}_{\text{rank}} 
    = \sum_{i<j}\max\!\Bigl(0,\;\text{scr}_{r,\sigma(i)} - \text{scr}_{r,\sigma(j)} + (j-i)\cdot\text{margin}\Bigr),
\end{equation}
where $i$ and $j$ are ground-truth stroke indices. This loss penalizes cases where a stroke that should appear earlier (smaller $i$) is assigned a higher or nearly equal score compared to a later stroke (larger $j$). The margin term requires a minimum separation proportional to the index gap, preventing strokes that are far apart in the sequence from collapsing into nearly identical scores.

The overall training objective is
\begin{equation}
    \mathcal{L}_{\text{total}} 
    = \lambda_m\,\mathcal{L}_{\text{match}} 
    + \lambda_r\,\mathcal{L}_{\text{rank}},
\end{equation}

\noindent\textbf{Composition.} Once trained, the predictor outputs both stroke parameters and their inferred order. Strokes are then generated by StrokeDiff and rendered sequentially layer by layer, producing a coherent painting with reduced overlap artifacts and more natural temporal structure.

\begin{figure}[t]
    \centering
    \includegraphics[width=0.8\linewidth]{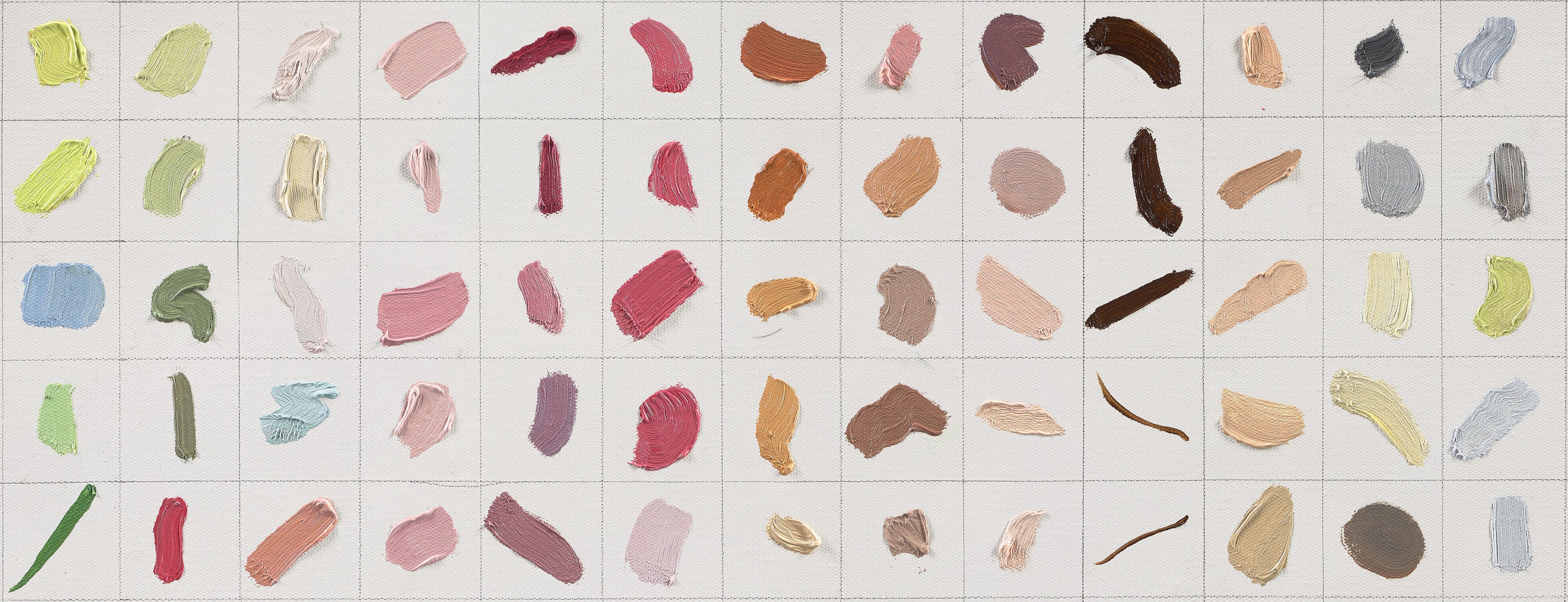}
    \caption{Examples of the collected strokes.}
    \label{fig:dataset}
    % \vspace{-3mm}  % 让正文更靠近图片
\end{figure}

\section{Experiments}

\subsection{Dataset}

To more closely mimic human painting behavior, we curated a dataset of 520 unique oil-painting strokes collected from a professional artist. Each stroke was drawn freely on a $20\times26$ grid canvas, as shown in Fig.~\ref{fig:dataset}, which illustrates the raw scanned sheet. Individual strokes were then cropped from the canvas and standardized to $295\times295$ pixels. We divided the dataset into 470 strokes for training and 50 for testing. To mitigate the limited sample size, we applied a series of preprocessing and augmentation steps, including background removal, color normalization, flipping, and rotation. This process expanded the dataset to 9,400 training images and 1,000 test image. The dataset is publicly available at \href{https://ieee-dataport.org/documents/artstroke}{DOI: 10.21227/t6bd-cs26} to support further research.

\subsection{Implementing details}
StrokeDiff is trained on a dataset of 9,400 augmented brushstroke images. For each training instance, we randomly sample 32 additional strokes from the dataset as visual priors, resulting in 300{,}800 training pairs. The model builds upon Stable Diffusion v1\_5 and generates outputs at a resolution of $512 \times 512$. We use the original noise schedule from Stable Diffusion, where $\bar{\alpha}_t = \prod_{i=0}^{t} (1 - \beta_i)$, with $\beta_t$ linearly interpolated between $\sqrt{0.00085}$ and $\sqrt{0.012}$. The prior upper bound $\Upsilon$ for $\eta$ is set to 0.5. Training is conducted using PyTorch 1.9.0. The model is first trained for 4 epochs without parameter conditioning, followed by 5 epochs of fine-tuning with conditional inputs. 

The stroke predictor operates on input canvases of size $32 \times 32$ and predicts up to 8 strokes per image. The matching loss weight vector $\lambda_m = [5, 10, 10]$ corresponds to the $L_1$, cosine, and decision loss terms, respectively. ${\lambda_r}=5$, and margin value is 0.125. The model was trained for 220 epochs.

\begin{table*}[t]
\centering
\begin{minipage}[t]{0.50\textwidth}
\centering
\caption{
Comparison of diffusion-based approaches for stroke generation, all trained on SD1.5. 
$n.$ and $r.$ denote the average \textbf{n}umber of closed regions per stroke and the \textbf{r}atio of region area to the full image, respectively. 
$\Delta$ indicates the change relative to the dataset-only baseline (FID = 47.3, computed between train and test; CRD (n./r.) = 1.09 (0.188)).}
\label{tab:comparison1}
\resizebox{\linewidth}{!}{ % ← 自动按宽度缩放
\begin{tabular}{lcccc}
\toprule
Method             & FID ↓ & $\Delta$ & CRD (n./r.) ↓ & $\Delta$ \\
\midrule
INS $b{=}0.3$\cite{chen2023importance}    & 273   & +225     & 48.51 (0.675) & +47. (+0.49) \\
INS $b{=}0.6$\cite{chen2023importance}    & 251   & +203     & 44.14 (0.732) & +43. (+0.54) \\
SD1.5\_LoRA\cite{hu2022lora}        & 285   & +237     & 88.13 (0.760) & +87. (+0.57) \\
\textbf{SmR (ours)} & \textbf{54} & \textbf{+6} & \textbf{1.24 (0.254)} & \textbf{+0.15 (+0.07)} \\
\bottomrule
\end{tabular}
}
\end{minipage}
\hfill
\begin{minipage}[t]{0.46\textwidth}
\centering
\captionof{table}{
Comparison with prior stroke generation methods for controllable generation, all trained on our dataset. \textsuperscript{$\dagger$} StrokeGAN and Stylized are conditioned on stroke parameters during training. RobPaint does not support conditional generation, so we evaluate it using its reconstruction outputs.}
\label{tab:comparison2}
\resizebox{\linewidth}{!}{
\begin{tabular}{l|c|c|c|c}
\toprule
\textbf{Model} & \textbf{Size} & \textbf{LPIPS}$\downarrow$ & \textbf{MSE}$\downarrow$ & \textbf{FID}$\downarrow$ \\
\midrule
StrokeGAN~\cite{wang2023stroke}        & $64 \times 64 \times 3$   & 0.285 & 0.064 & 245 \\
Stylized~\cite{zou2021stylized}        & $128 \times 128 \times 3$ & 0.242 & 0.036 & 273 \\
RobPaint~\cite{bidgoli2020artistic}$^\dagger$ & $64 \times 32 \times 1$   & 0.234 & 0.085 & 322 \\
\textbf{StrokeDiff}                    & \textbf{$512 \times 512 \times 3$} & \textbf{0.232} & \textbf{0.031} & \textbf{52} \\
\bottomrule
\end{tabular}
}
\end{minipage}

\end{table*}

\begin{figure*}[t]
    \centering
    \includegraphics[width=1\linewidth]{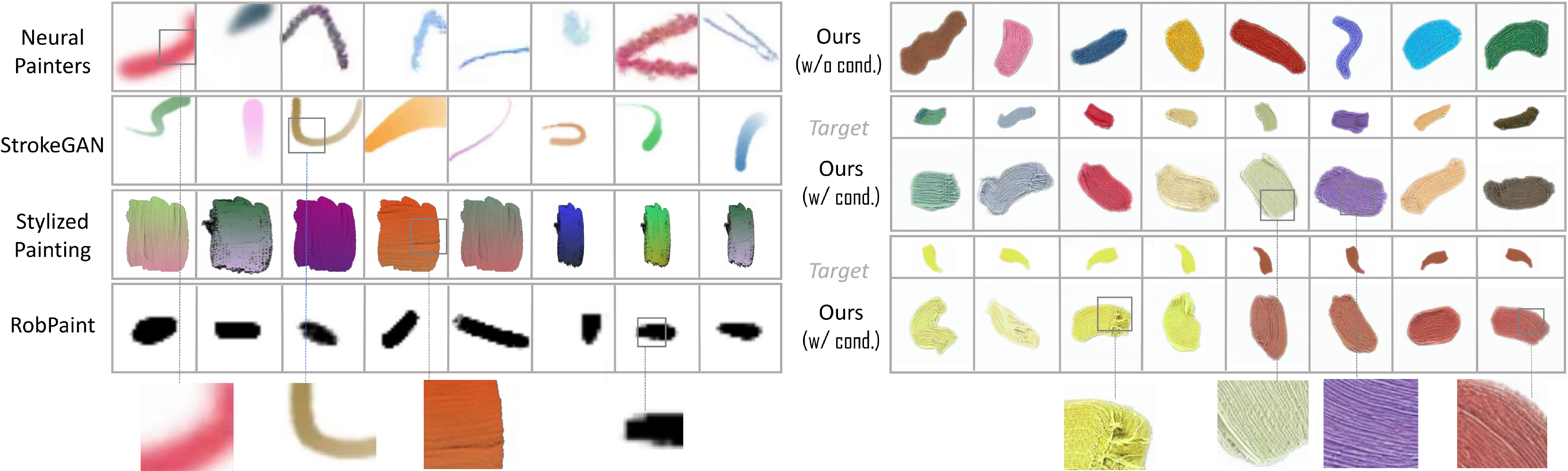}
    % \vspace{-5mm}
    \caption{Qualitative comparison with other approaches: 1) Neural Painters \cite{nakano2019neural}, using VAE and GAN trained on synthetic MyPaint strokes; 2) StrokeGAN \cite{wang2023stroke}, a GAN trained on fourth-order Bézier curves; 3) Template strokes \cite{zou2021stylized, liu2021paint, tong2022im2oil,song2024processpainter}, taken from a widely used set of shapes commonly adopted in SBR; and 4) RobPaint \cite{bidgoli2020artistic}, a VAE trained with human grayscale strokes. Our method generates clearer strokes while preserving brush textures.}
    \label{fig:comparison}
    % \vspace{-3mm}
\end{figure*}

\subsection{Impact of Smooth Regularization on Stroke Generation}
\subsubsection{Evaluating generalization in few-shot diffusion} Table~\ref{tab:comparison1} reports quantitative results comparing SmR with representative low-resource training strategies for unconditional stroke generation, including noise schedule adjustment and LoRA-based fine-tuning. Evaluation is conducted using FID \cite{heusel2017gans}, as well as a structure-sensitive metric we introduce, Closed Region Detection (CRD). The motivation for CRD is that failures in stroke generation often manifest as either fragmented artifacts resembling multiple small brush-like patches or, conversely, as large blobs where strokes merge into indistinct masses. To capture these structural failures, CRD measures (i) the number of closed regions per image and (ii) the proportion of stroke area relative to the canvas. Both are computed via connected component analysis in OpenCV. 
% A high number of small closed regions indicates over-fragmentation, while a single oversized region suggests collapsed or over-smoothed outputs. 
For noise scheduling, we adapt \textit{strategy 2} from INS~\cite{chen2023importance}, modifying the scaling factor $b$ in the forward noise term to mitigate its counteracting effect. Setting $b = 1$ corresponds to the default schedule. We evaluate $b$ values of $0.3 \text{ and } 0.6$. Although LoRA~\cite{hu2022lora} was originally proposed for efficient fine-tuning, we include it as a strong low-data baseline due to its favorable parameter efficiency. Despite using only 470 hand-drawn strokes, SmR achieves a significantly lower FID. Both INS-based noise scheduling and LoRA fail to prevent structure collapse or mode proliferation: the number of detected regions is significantly overestimated, and FID remains high (qualitative examples can be seen in Fig.~\ref{fig:ablation} columns 2–4).

\subsubsection{Quantitative comparison with stroke generation methods}\label{sec:results_comp}We further evaluate SmR in the context of controllable stroke generation, comparing against prior work including StrokeGAN~\cite{wang2023stroke}, Stylized~\cite{zou2021stylized}, and RobPaint~\cite{bidgoli2020artistic}. Table~\ref{tab:comparison2} reports LPIPS~\cite{zhang2018unreasonable}, MSE, and FID scores, where our model applied with SmR outperforms all methods across all metrics. We note that stroke data, unlike natural images, often exhibits lower perceptual and pixel-level variance, which limits the sensitivity of LPIPS and MSE in distinguishing structural quality. We found FID remains more discriminative in capturing distributional mismatch in the experiments, and SmR achieves the lowest FID, indicating that its outputs better align with the training data distribution. Qualitative results are shown in Appendix.

\begin{figure*}[t]
    \centering
    \includegraphics[width=1\linewidth]{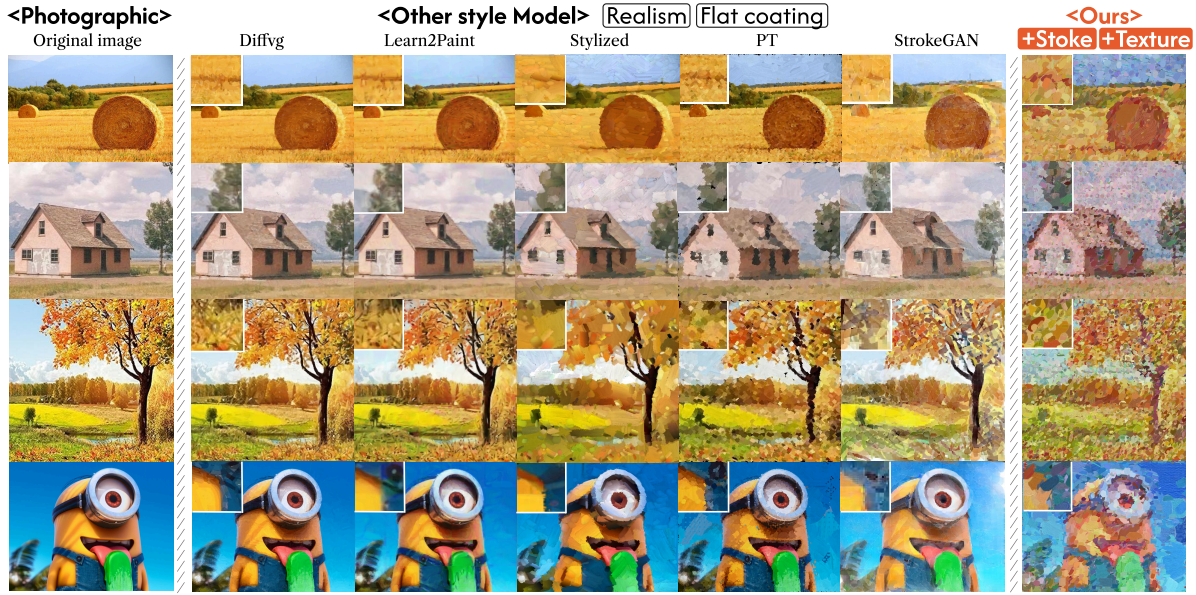}
    \caption{Qualitative comparison of paintings produced by different stroke-based rendering pipelines: Diffvg~\cite{li2020differentiable}, Learn2Paint~\cite{huang2019learning}, Stylized~\cite{zou2021stylized}, PT~\cite{liu2021paint}, and StrokeGAN~\cite{wang2023stroke}. The figure illustrates how different stroke generation strategies affect texture appearance, stroke irregularity, and layering behavior in the final compositions. Zoom in to see differences.
}
    \label{fig:paintings}
    % \vspace{-3mm}
\end{figure*}

\subsubsection{Qualitative comparison of strokes in other methods}
Fig.~\ref{fig:comparison} shows qualitative results from several representative stroke generation models. Neural Painters \cite{nakano2019neural} and StrokeGAN~\cite{wang2023stroke} produce smooth outputs but show limited texture fidelity. Template-based approaches exhibit high pixel accuracy but constrained variability, while RobPaint \cite{bidgoli2020artistic} generates more organic strokes but lacks color expressiveness due to grayscale training. In comparison, StrokeDiff demonstrates both unconditional and conditional generation capabilities: the unconditional one-shot setting outputs feature diverse, textured strokes with clear structural integrity. When conditioned on specific parameters, the model accurately matches the targets in both shape and appearance. The final row shows augmented targets used in training, highlighting the model’s capacity to capture a wide range of brushstroke styles.

\subsection{StrokeDiff in Downstream Painting Tasks}

We next evaluate the rendering stage described in Section~\ref{sec:pipeline}, focusing on the quality of complete paintings. We compare our method against representative approaches, including Diffvg~\cite{li2020differentiable}, Learn2Paint~\cite{huang2019learning}, PaintTransformer (PT)~\cite{liu2021paint}, Stylized~\cite{zou2021stylized}, and StrokeGAN~\cite{wang2023stroke}. These methods span differentiable vector graphics, reinforcement learning, feed-forward prediction, and GAN-based stroke generation.

\subsubsection{Qualitative evaluation of paintings}
\begin{figure}[t]
    \centering
    \includegraphics[width=0.72\linewidth]{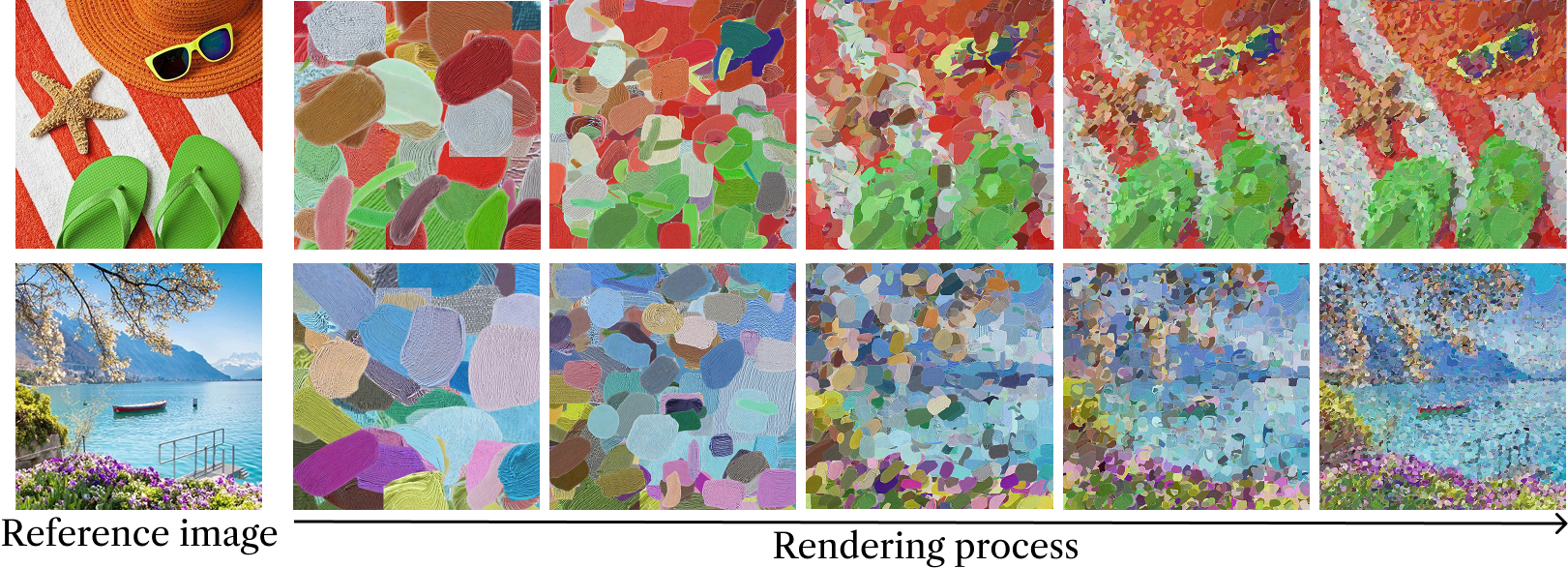}
    \caption{Intermediate canvases in the rendering process.}
    \label{fig:progress}
\end{figure}

Fig.~\ref{fig:paintings} compares the painting results produced by different stroke-based rendering approaches. Diffvg and Learn2Paint focus on accurate pixel-level reconstruction, resulting in images that closely follow the reference but rely on relatively uniform and geometry-driven stroke patterns. Stylized and StrokeGAN introduce higher shape variability, yet their strokes often exhibit simplified textures or less consistent layering behavior. Our method adopts a different emphasis by modeling strokes as texture-rich and irregular visual primitives. As a result, the rendered paintings show more pronounced stroke texture, clearer layering structure, and a visual appearance that aligns more closely with the characteristics of oil painting. Fig.~\ref{fig:progress} further illustrates how such stroke-level properties accumulate during the sequential rendering process.

\subsubsection{Quantitative evaluation of paintings}
Table~\ref{tab:clip_score} reports semantic alignment measured by CLIP score. We evaluate each method on five images using four descriptive captions: \emph{“an oil painting made by a human”, “a brushstroke oil painting”, “a textured oil painting”}, and \emph{“a stylized oil painting with strong brushstroke texture made by a human”}. Results show that our approach reaches a similar score to Stylized on the first caption. On the remaining three captions, which emphasize brushstroke texture and stylization, our method consistently surpasses all alternatives. These improvements suggest that StrokeDiff better aligns with human-oriented descriptions of oil painting, especially in aspects related to stroke quality and artistic expression.

\begin{table*}[t]
    \centering
    % \footnotesize

    % ====== 左表 CLIP Score ======
    \begin{minipage}[t]{0.51\textwidth}
        \centering
        \caption{CLIP score comparison across different methods. Ori. denotes reference images.}
        \label{tab:clip_score}
        \renewcommand{\arraystretch}{1.1}
        \setlength{\tabcolsep}{4pt}
        \resizebox{\linewidth}{!}{
        \begin{tabular}{cl|ccccccc}
            \multicolumn{2}{c|}{\textbf{Method}}  
            & \textbf{\textcolor{gray}{Ori.}} & \textbf{Diffvg} & \textbf{L2P} & \textbf{Styl.} 
            & \textbf{PT} & \textbf{SGAN} & \textbf{Ours} \\ 
            \midrule
        \multirow{4}{*}{\shortstack{\textbf{CLIP} \\ \textbf{Score}}}
        & cap.1 & \textcolor{gray}{0.125} & 0.1704 & 0.1879 & \cellcolor{gray!10}0.1981 & 0.1751 & 0.1886 & 0.1951 \\ 
        & cap.2 & \textcolor{gray}{0.117} & 0.1757 & 0.2004 & 0.2059 & 0.1896 & 0.2019 & \cellcolor{gray!10}0.2089 \\ 
        & cap.3 & \textcolor{gray}{0.106} & 0.1710 & 0.1970 & 0.2079 & 0.1901 & 0.1960 & \cellcolor{gray!10}0.2226 \\ 
        & cap.4 & \textcolor{gray}{0.131} & 0.1950 & 0.2254 & 0.2294 & 0.2079 & 0.2097 & \cellcolor{gray!10}0.2363 \\  
        \end{tabular}
        }
    \end{minipage}
    \hfill
    % ====== 右表 Human Rating ======
    \begin{minipage}[t]{0.45\textwidth}
        \centering
        \caption{Human evaluation on painting quality (1–5 scale).}
        \label{tab:human_painting}
        \renewcommand{\arraystretch}{1.0}
        \setlength{\tabcolsep}{5pt}
        \resizebox{\linewidth}{!}{
        \begin{tabular}{l|cccccc}
            & Diffvg & L2P & Stylized & PT & S.GAN & \textbf{Ours} \\
            \midrule
                Style    & 3.21 & 3.09 & 3.35 & 3.18 & 3.26 & \textbf{3.38} \\
                Aesthetic      & 2.87 & 2.50 & 3.00 & 2.76 & 3.11 & \textbf{3.38} \\
                Texture    & 2.84 & 2.39 & 3.18 & 2.79 & 2.94 & \textbf{3.50} \\
                Content & \textbf{4.35} & 4.12 & 3.37 & 3.48 & 3.39 & 2.59 \\
                % \midrule
                Total      & \textbf{3.32} & 3.03 & 3.24 & 3.05 & 3.17 & 3.21 \\
        \end{tabular}
        }
    \end{minipage}
    % \vspace{-10mm}
\end{table*}

\subsubsection{User Study}
\label{sec:user_study}
Following \cite{wang2024learning}, we conducted a human evaluation with 51 participants, including 26 with formal art backgrounds and 25 without. Participants rated each method on four criteria: \emph{oil painting style}, \emph{aesthetic value}, \emph{texture}, and \emph{content retention}. As shown in Table~\ref{tab:human_painting}, our approach earned the highest ratings for style, aesthetics, and texture. However, its emphasis on artistic expression led to a lower content retention score, suggesting a trade-off between fidelity and stylization. The questionnaire is provided in Appendix.

\begin{figure*}[t]
  \centering
  \subfloat[\footnotesize Stochastic vs. Deterministic.\label{fig:sto}]{
    \includegraphics[width=0.33\textwidth]{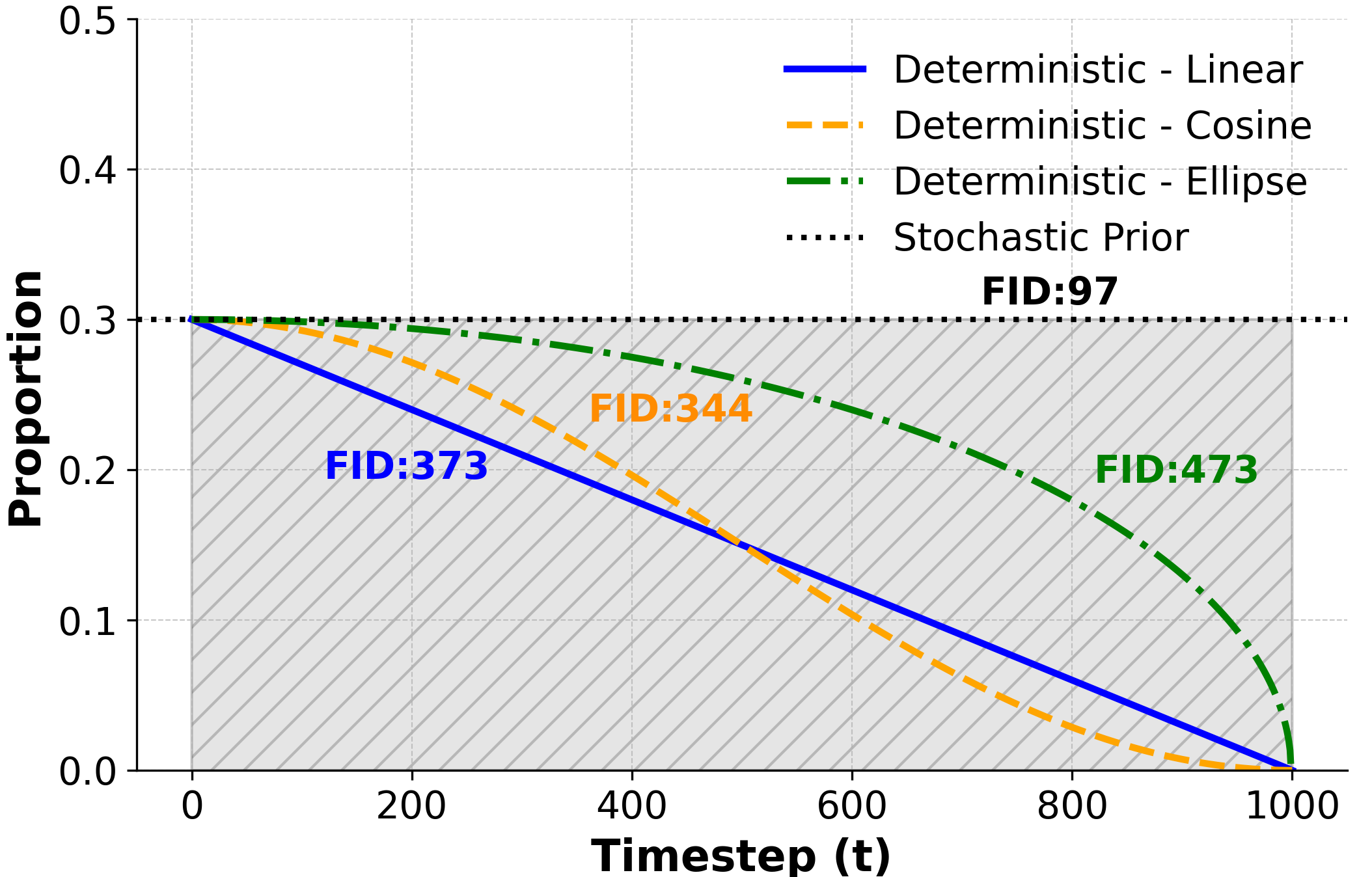}
  }
  \subfloat[\footnotesize Evaluation on $\Upsilon$ values.\label{fig:eta_value}]{
    \includegraphics[width=0.33\textwidth]{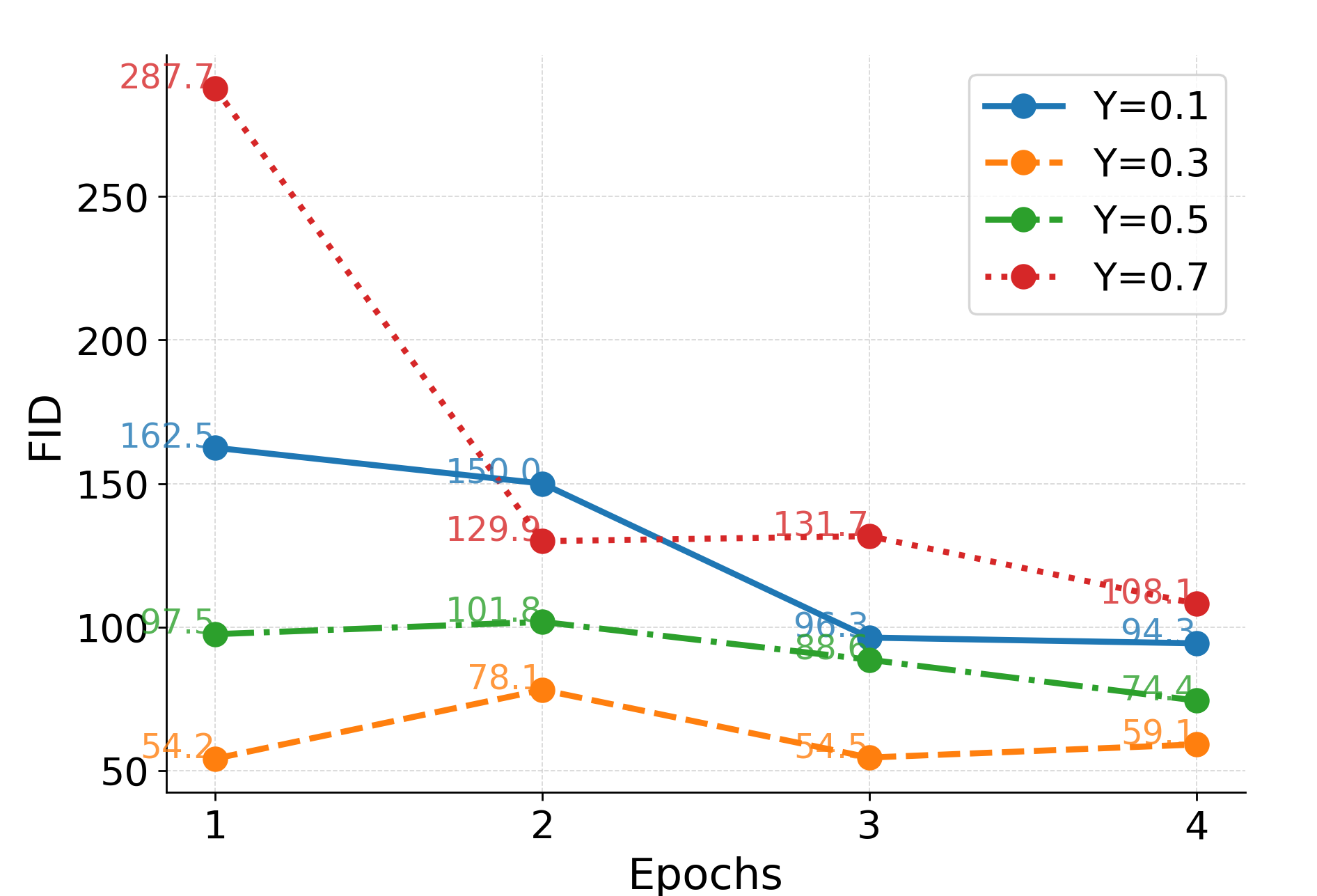}
  }
  \subfloat[\footnotesize Signal-to-Noise change.\label{fig:snr}]{
    \includegraphics[width=0.33\textwidth]{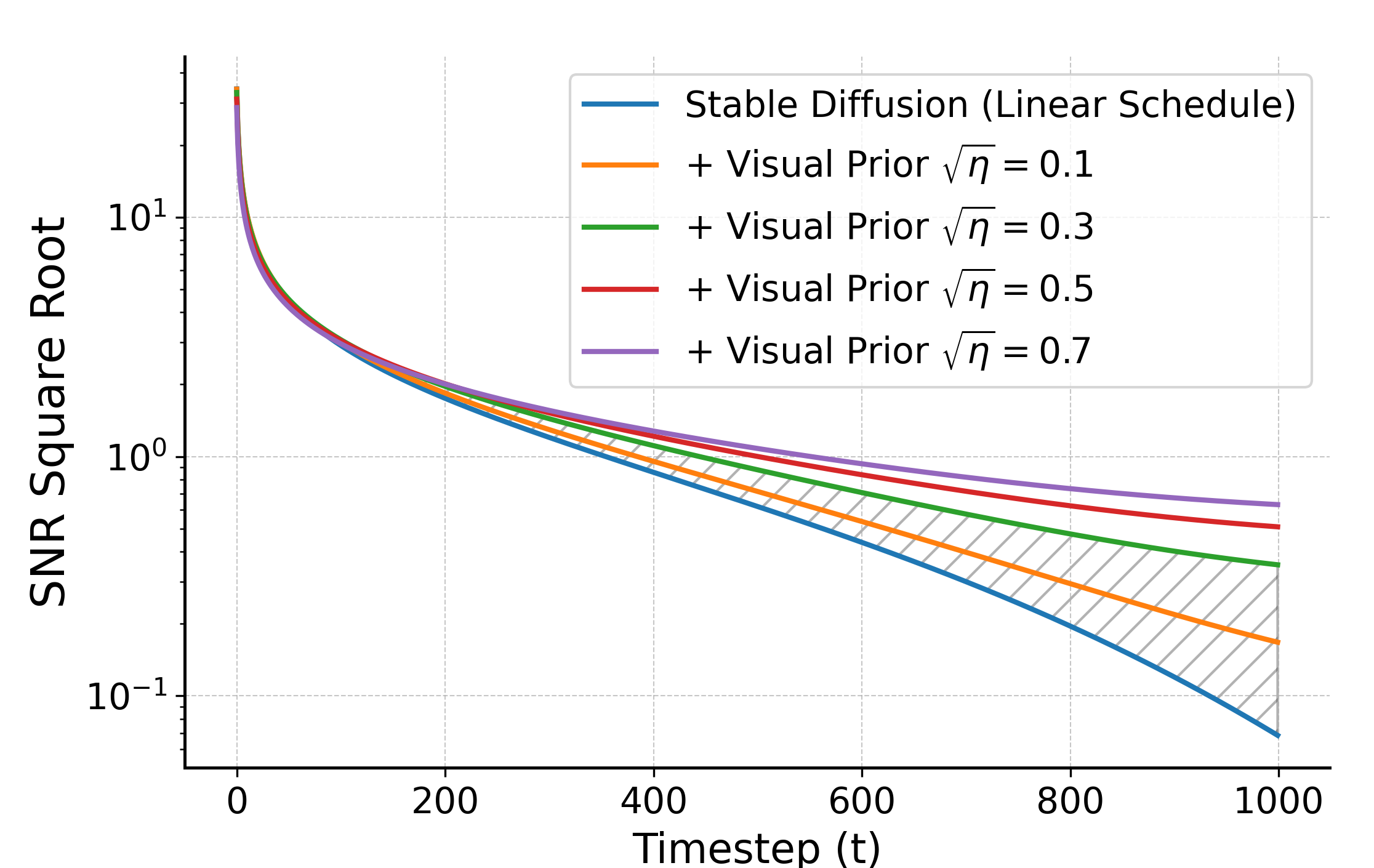}
  }
  \caption{Ablation results. (a) Injection trend of stochastic and deterministic prior; (b) FID across training epochs for different $\Upsilon$ values; (c) SNR changes. We take the square root of standard SNR.}
  \label{fig:all_three}
\end{figure*}

\begin{table}[t]
    % \vspace{-3mm}
    \centering
    \footnotesize
    \caption{Evaluation on different numbers of target-prior pairs for each data sample. The left and right groups correspond to cases where $\Upsilon$ is set to 0.3 and 0.5.}
    \vspace{2mm}
    \resizebox{0.55\columnwidth}{!}{
        \begin{tabular}{ccc>{\columncolor{gray!15}}c|ccc}  
            \textbf{\# Multiple Pairs $\rightarrow$} & \textbf{10} & \textbf{20} & \textbf{32} & \textbf{10} & \textbf{20} & \textbf{32}\\ 
            \midrule
            FID & 148.6 & 91.6 & 54.2 & 108 & 97.5 & 74.4\\
        \end{tabular}
    }
    \label{tab:pairs}
    % \vspace{-3mm}
\end{table}

\subsection{Ablation studies}
\subsubsection{Analyzing the Role of Visual Prior Injection} 
\paragraph{Evaluation on stochastic visual prior}We compare our stochastic prior injection strategy with deterministic alternatives. In our method, the injection factor is randomly sampled and remains step-independent, while deterministic priors follow a fixed schedule across time—analogous to noise schedules in standard diffusion models (e.g., linear, cosine, or scaled cosine schedules), commonly used for controlling the diffusion strength via predefined $\alpha_t$ or $\beta_t$ curves. We test several such deterministic variants, including linear, cosine, and ellipse-shaped prior weights. As shown in Fig. \ref{fig:sto}, our approach achieves a significantly lower FID,  as it prevents the model from propagating fixed information, enabling the model to better adapt to stroke variations and learn a broader sample distribution. The \textit{gray-shaded} area in the figure represents the range of possible prior values introduced by the stochastic method, highlighting that it is no longer a fixed trajectory.

% We compare stochastic and deterministic visual priors, where the stochastic prior samples a variable injection factor (bounded by $\Upsilon$) at each timestep, while deterministic methods apply a fixed prior. We tested common deterministic strategies, including linear cosine, and ellipse scheduling. 

\paragraph{Evaluation on prior upper bound}We evaluated different values of the prior upper bound $\Upsilon$. As shown in Fig. \ref{fig:eta_value}, the model can converge across four tested values, with clear early-stage convergence at $\Upsilon=$0.3 and 0.5, particularly at 0.3, where the model stabilized within the first epoch. In our other experiments, we found that $\Upsilon=0.5$ performs comparably to $=0.3$ after conditioning training, but was more robust in low-data scenarios. Therefore, setting $\Upsilon=$0.3 accelerates training when data is sufficient, while a higher upper bound provides stronger guidance in low-data settings. Table \ref{tab:pairs} shows results for different numbers of priors matches per sample, where larger $\Upsilon$ is beneficial when fewer pairs are available. We did not test beyond 32 pairs, as this was already sufficient to produce high-quality strokes.

\paragraph{Effect on SNR dynamics}
\label{sec:NS}
Although our method does not explicitly perform noise scheduling, injecting visual priors during the forward process inherently modifies the signal-to-noise ratio (SNR). This connection offers a complementary perspective on how SmR improves training. Fig.\ref{fig:snr} illustrates the expected SNR trajectory when $\sqrt{\eta} = 0.3$ is applied uniformly across timesteps, treating the prior as part of the signal (green curve). In practice, due to stochastic injection, the actual SNR fluctuates within the shaded band, consistently higher than that of standard Stable Diffusion. 

\begin{figure*}[t]
    % \vspace{-3mm}
    \centering
    \includegraphics[width=1\linewidth]{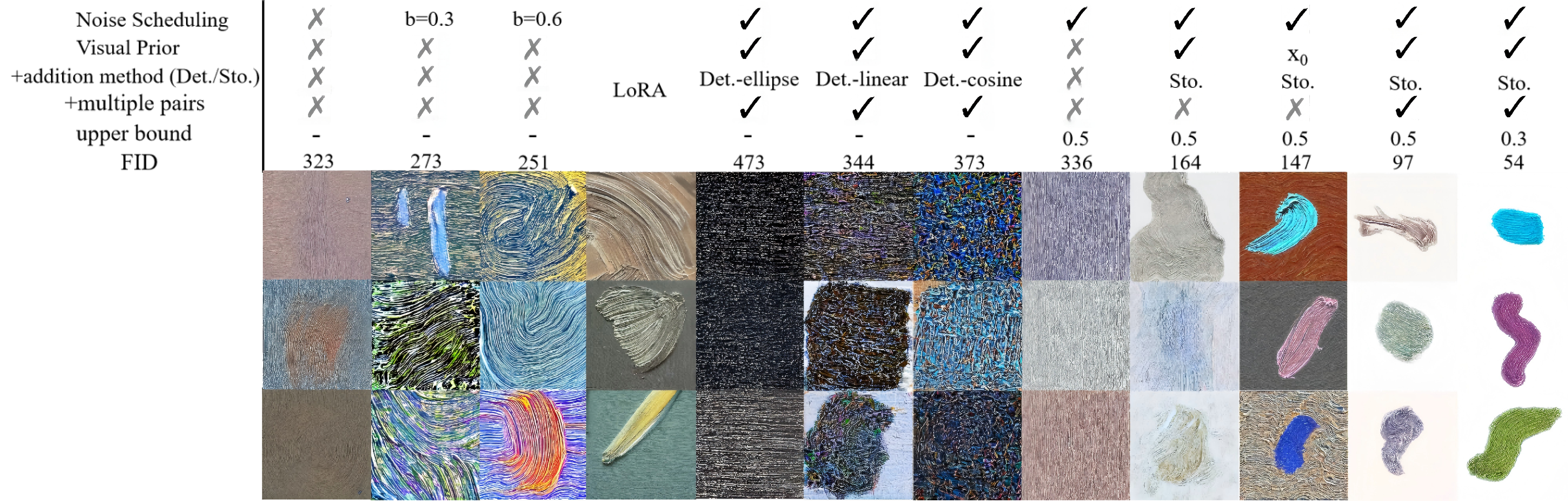}
    \caption{Example results from ablation experiments. Since the default prior-sample pairs are 32, all "w/o multi-pairs" settings were trained for 32 epochs to match the total number of updates, while other settings were trained for 1 epoch. $Det.$ and $Sto.$ denote deterministic and stochastic prior injection methods. Although LoRA is not part of our ablation, we include its results in the fourth column for qualitative comparison.}
    \label{fig:ablation}
    % \vspace{-2mm}  % 让正文更靠近图片
\end{figure*}

\paragraph{Augmentation effect}To isolate the augmentation effect of SmR, we further test a degenerate case where the prior $x_s$ is replaced with $x_0$ (ground truth). As shown in Fig.~\ref{fig:ablation} (third-to-last column), even without introducing new information, this variant significantly reduces FID, suggesting that SmR acts as an effective training-time regularizer beyond data augmentation. Exploring this property in other domains is an interesting direction for future work.

\begin{figure}[t]
    \centering
    % 第一个子图：占单栏48%宽度（预留间隙），置顶对齐
    \begin{minipage}[t]{0.35\linewidth}
        \centering
        \includegraphics[width=1\linewidth]{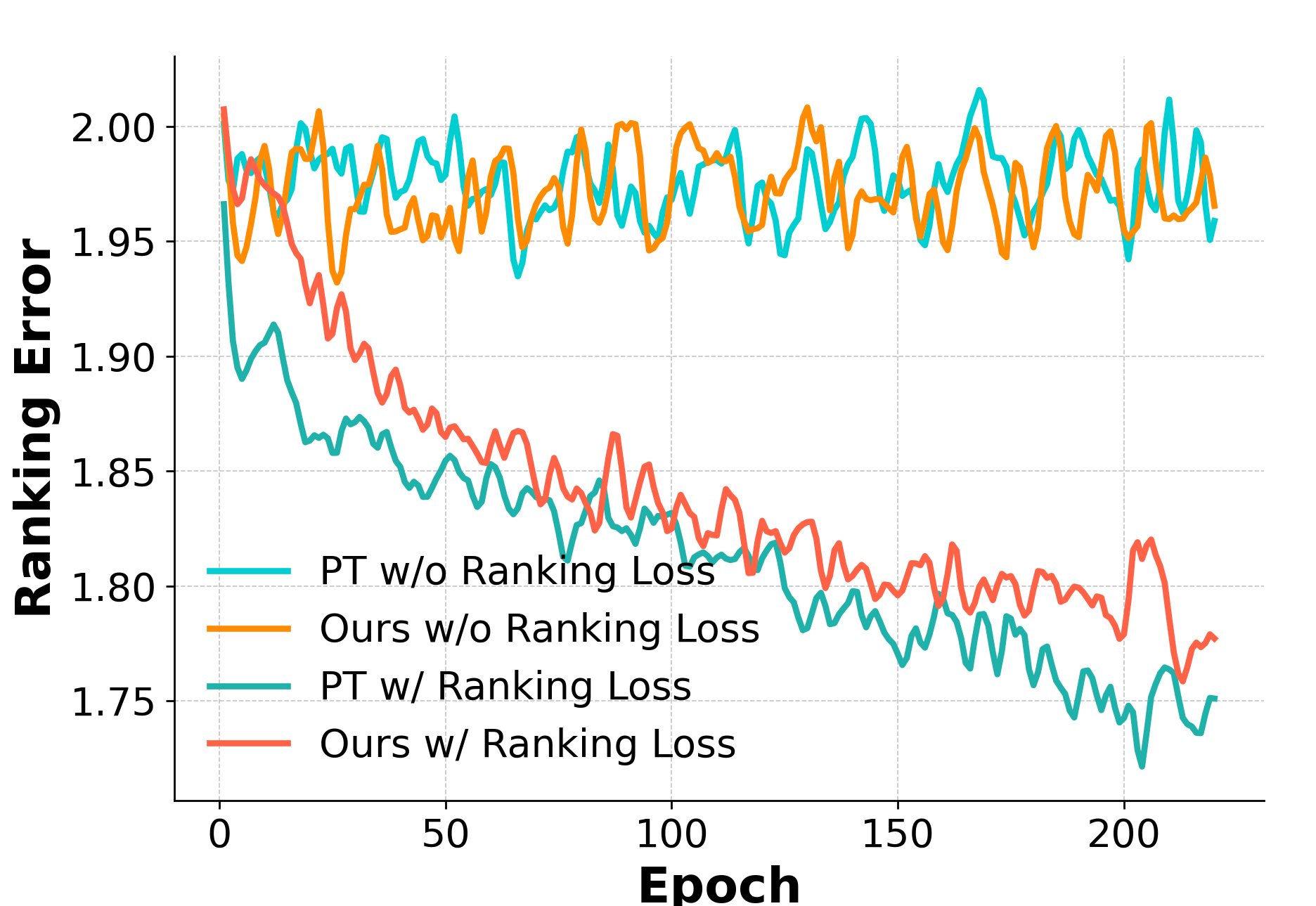}
        \caption{Rank error w/ and w/o Ranking Loss. For the method w/o Ranking Loss, $src_r$ is replaced by predicted index $\times$ margin.}
        \label{fig:rank}
    \end{minipage}
    \hfill  % 自动填充两图之间的空白（均匀分布）
    % 第二个子图：与左侧等宽，置顶对齐
    \begin{minipage}[t]{0.64\linewidth}
        \centering
        \includegraphics[width=1\linewidth]{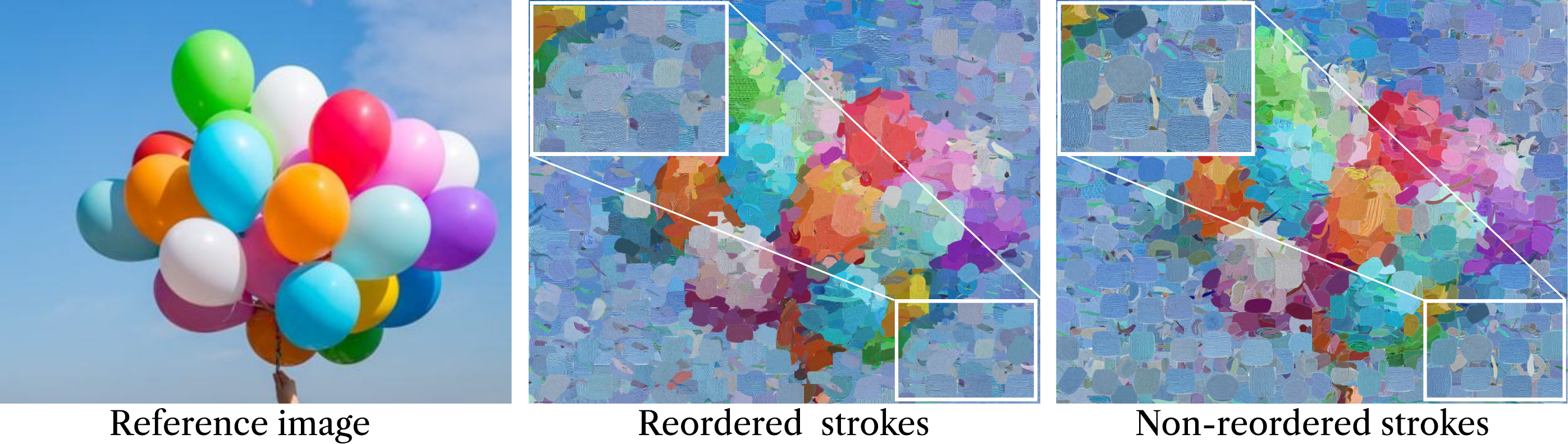}
        \caption{Paints w/ and w/o reordered strokes using predicted rank scores.}
        \label{fig:reorder}
    \end{minipage}
\end{figure}

\subsubsection{Ablation on ranking loss}
We assessed the stroke predictor’s ability to maintain a consistent rendering order by measuring sequence error between predicted and ground-truth stroke orders, with and without Ranking Loss. As shown in Fig.~\ref{fig:rank}, without the Ranking Loss, the model’s set-based prediction approach leads to a completely unordered sequence, which affects the overall coherence of the painting. With the Ranking Loss, the stroke order progressively aligns with the correct sequence. We also tested this approach on PT \cite{liu2021paint} and observed similar improvements. Fig.\ref{fig:reorder} further illustrates that without reordering, strokes exhibit noticeable layering artifacts, whereas reordering produces a more cohesive blending.

\subsection{Generality Beyond Stroke Generation}
\label{sec:inpainting}
To examine whether Smooth Regularization can benefit domains other than stroke synthesis, we conducted a preliminary test on the inpainting task Paint-by-Example~\cite{yang2023paint}, which provides a complete training and evaluation framework. Following the same training protocol, we applied a degenerate form of SmR where the visual prior was set to the input $x_0$. Both the baseline and the SmR-augmented models were trained on subsets of the Open-Images~\cite{kuznetsova2020open} dataset, using either 8\% or 1\% of the training data. Table~\ref{tab:openimages} summarizes the results. While performance at the 8\% data level is broadly comparable, SmR yields modest gains in the extremely low-data regime (1\%), with lower FID and higher quality scores. These preliminary results suggest that SmR may act as a useful training-time regularizer beyond stroke generation, although broader validation across diverse tasks remains an open direction for future work.

\begin{table}[t]
\centering
\caption{Results on Paint-by-Example with subsets of Open-Images.}
\label{tab:openimages}
% 核心紧凑化设置（单栏生效，双栏不影响）
\footnotesize  % 比small更小，进一步压缩
\setlength{\tabcolsep}{9pt}  % 列间距压到最小（保证可读性）
\setlength{\extrarowheight}{1pt}  % 轻微增加行高，避免文字挤在一起

% 固定表格总宽度（单栏：0.7\linewidth，双栏：\linewidth，自动适配）
\begin{tabular}{@{} lccc|ccc @{}}
\toprule
& \multicolumn{3}{c}{\textbf{8\% Open-Images}} & \multicolumn{3}{c}{\textbf{1\% Open-Images}} \\[0.3ex]
Method & FID ↓ & QS ↑ & CLIP ↑ & FID ↓ & QS ↑ & CLIP ↑ \\
\midrule
Paint-by-Example & 11.667 & 73.254 & \textbf{74.120} & 27.421 & 27.473 & \textbf{58.827} \\
+ SmR ($x_0$)    & \textbf{11.467} & \textbf{73.981} & 73.544 & \textbf{26.858} & \textbf{31.148} & 57.179 \\
\bottomrule
\end{tabular}
\end{table}

\section{Discussion}
\subsection{Potential application senarios}
Beyond the scope of our current experiments, it is worth considering how stroke-level generation could be applied in related domains.

\textbf{Robotic painting.} One natural direction is robotic painting, where diffusion-based strokes may be mapped to robotic arm trajectories in order to reproduce brush-like appearances with greater fidelity. Recent systems such as FRIDA~\cite{schaldenbrand2022frida} and COFRIDA~\cite{schaldenbrand2024cofrida} have already shown the feasibility of combining generative models with robotic painting, and stroke-level synthesis could further enrich this process by providing more diverse and textured visual units.

\textbf{Creative support tools.} CSTs are interactive systems that assist the creative process by providing real-time, controllable enhancements while keeping the user in primary creative control. In this context, the texture expressiveness of our strokes could be extended to serve as rendering units embedded in digital painting toolchains, such as styluses or touchscreen brushes. This idea is related to Diffusion Texture Painting~\cite{hu2024diffusion}, which demonstrated how diffusion models can enrich user input with textured feedback. StrokeDiff differs in that it focuses on stroke-level synthesis, and could therefore be explored in future as a complementary component for producing painterly, layered brush effects in educational or creative software.

\textbf{2.5D printing.} Advances in 2.5D and relief printing technologies (e.g., Canon’s surface-relief printing) have made it possible to reproduce tactile qualities in artworks. However, existing pipelines typically rely on heuristic mappings from RGB values to height maps, which may fail to capture fine brush textures. Because our strokes explicitly encode layered and textured appearances, they could be converted into printable height fields that preserve the physical feel of oil painting. Such an application would expand the use of stroke synthesis from visual rendering to tangible reproduction.

\subsection{Evaluation Challenges in AI-Based Painting}
In our user study, participants were asked to rate “From the photograph to the painting, how well is the content applied in the artwork?”. Compared with style, aesthetics, and texture, the scores for this content-related question were noticeably lower. This raises an important question: does a lower content score necessarily imply weaker performance, or does it instead reflect the tension between abstraction and fidelity that is central to non-photorealistic rendering? In our case, the generated strokes do not alter object presence or layout, but reinterpret the same content with varying degrees of abstraction, which makes the notion of “content preservation” itself somewhat ambiguous. From one viewpoint, reduced fidelity is a shortcoming; from another, it is an expected property of artistic stylization. This ambiguity points to a broader issue in the evaluation of AI-based painting systems. Much of the existing work on aesthetic assessment \cite{jin2024apddv2} or art criticism \cite{bin2024gallerygpt} focuses on finished artworks, which often focus on finished masterpieces or historical paintings, differing substantially from process-oriented generative painting. Consequently, there remains a lack of comprehensive benchmarks that can assess painting systems across multiple dimensions, such as semantic coherence, abstraction level, stroke expressiveness, and stylistic intent, depending on task requirements. Widely-used image metrics, including FID or CLIP-based similarity, primarily capture distributional alignment or coarse semantic correspondence, but are insensitive to non-rigid, perceptual qualities central to artistic rendering. Recent evaluation protocols for generative tasks have begun to incorporate large multimodal language models as flexible judges, for example through GPT-based or VQA-style assessments in benchmarks such as T2I-CompBench++ \cite{huang2025t2i}, as well as through general-purpose VLMs including the LLaVA family ~\cite{liu2023visual} and Qwen2-VL ~\cite{wang2024qwen2}. These models enable evaluation along qualitative dimensions that are difficult to formalize numerically. While still imperfect, MLLM-based evaluation offers a promising complementary direction for assessing AI painting systems beyond traditional pixel- or embedding-level metrics.

\section{Limitations and future work}

\subsection{Rendering pipeline}
In this work we adopted a PaintTransformer-style predictor as the downstream rendering pipeline, primarily because its feed-forward design is computationally efficient. However, this design choice also inherits limitations that are common to similar approaches such as Learn2Paint and StrokeGAN. In particular, the grid-based layer-wise rendering strategy divides the canvas into $2^k\times2^k$ patches at $k$-th layer, leading to strokes that are locally constrained and often uniform in size. This behavior differs from human painting practices, where artists naturally vary stroke length, orientation, and continuity depending on the region and intended effect. An alternative direction would be to integrate StrokeDiff into frameworks such as Diffvg, where stroke placement is jointly optimized over the canvas. In principle, such integration could allow longer strokes that align with object contours and reduce the grid artifact, although it would require adjustments in stroke parameterization and pipeline integration.

\subsection{Style diversity and generalization}
StrokeDiff is particularly suited to texture-rich colored painting styles such as oil painting, where irregular boundaries and layered pigments are central to the visual effect. Other domains, such as sketch strokes, have very different visual characteristics: they are typically sparse, monochromatic, and lack continuous texture, which suggests that vector-domain modeling with rasterization might be a more natural choice than diffusion-based texture synthesis. For other colored media, including watercolor and ink wash, we have not yet attempted generalization. The main limitation here is the lack of domain-specific datasets, which makes it difficult to capture phenomena such as transparency, pigment blending, and negative-space effects. To explore this direction, we are currently constructing a dataset of 1,008 ink-style brushstrokes. This resource will allow us to examine stroke generation in a medium with different physical properties and may inform future adaptations of stroke parameterization and rendering rules.

\subsection{Generalization of SmR}
Although we included a preliminary experiment in Section \ref{sec:inpainting} using an inpainting framework, this alone is not sufficient to demonstrate that SmR generalizes across domains. Whether it can serve as a general-purpose regularization strategy remains to be established through broader validation. Nevertheless, for tasks where data collection is difficult or where the goal is to generate fundamental primitives as building blocks, SmR could have potential value and merits further investigation.

\section{Conclusion}
This work explored the problem of brushstroke generation under limited data and introduced StrokeDiff with Smooth Regularization as a data-efficient diffusion framework. The results show that stroke-level modeling can be both structurally diverse and stylistically expressive, and that such primitives can be composed into convincing paintings. More broadly, this study illustrates how generative models can contribute to non-photorealistic rendering when adapted to domain-specific primitives. Future work may extend this direction to other artistic media, and explore more comprehensive evaluation protocols that account for both abstraction and fidelity.

\begin{acks}
This work is supported by the Research Grants Council of Hong Kong under the Theme-based Research Scheme (T45-205/21-N).
\end{acks}

\bibliographystyle{ACM-Reference-Format}
\bibliography{TOMM}

\appendix

Online Appendix

\section*{Formula derivation of SmR}
\subsection*{Proof of the Second Zonklar Equation}
The forward process in diffusion models, such as DDPMs, describes how data transitions over time by progressively adding noise. The transitions between time steps can be represented as:

\begin{align}
x_t &= \sqrt{\alpha} x_{t-1} + \sqrt{1-\alpha} \, \epsilon_{t-1}, \\
x_{t-1} &= \sqrt{\alpha} x_{t-2} + \sqrt{1-\alpha} \, \epsilon_{t-2}, \\
&\vdots \\
x_2 &= \sqrt{\alpha} x_1 + \sqrt{1-\alpha} \, \epsilon_1, \\
x_1 &= \sqrt{\alpha} x_0 + \sqrt{1-\alpha} \, \epsilon_0.
\end{align}

We introduce an additional visual prior $x_s$ into various stages of the diffusion process, allowing semantic information to guide the process, making the modified forward process partially dependent on this visual prior:

\begin{align}
x_t' &= x_t + \sqrt{1-\bar{\alpha}_t} \, \sqrt{\eta} x_s - \sqrt{1-\bar{\alpha}_t} \, \sqrt{\eta} \cdot \epsilon_t^*, \\
x_{t-1}' &= x_{t-1} + \sqrt{1-\bar{\alpha}_{t-1}} \, \sqrt{\eta} x_s - \sqrt{1-\bar{\alpha}_{t-1}} \, \sqrt{\eta} \cdot \epsilon_{t-1}^*, \\
&\vdots
\end{align}

Where $\epsilon^*\sim\mathcal{N}(0,I)$ is independently sampled noise, and $\sqrt{\eta}\sim Uni[0,0.5)$ is a scaling factor. $x_s$ is added directly at each timestep, but it does not evolve or propagate iteratively through the sequence. 
Therefore, $x_t'$ is derived from $x_t$ and is implicitly related to $x_{t-1}'$.

\subsection*{Proof: The Sequence \ensuremath{X = \left\{ x_0, x_1^\prime, x_2^\prime, \ldots, x_t^\prime \right\}} is Markovian}
To utilize Bayes' theorem for calculating the new sampling distribution, \(q(x_t' | x_{t-1}', x_0)\) and \(q(x_t' | x_{t-1}')\) must be equivalent. Therefore, it is necessary to prove that the sequence \(X = \{x_0, x_1', x_2', \ldots, x_t'\}\) is Markovian.

The sequence dependencies are defined as:
\begin{align}
    x_{t-1} \to x_{t-1}' = f(x_{t-1}, \epsilon_{t-1}^*, x_s), \quad  
    x_{t-1} \to x_t = g(x_{t-1}, \epsilon_{t-1}), \quad
    x_t \to x_t' = h(x_t, \epsilon_t^*, x_s),
\end{align}
$x_s$ is provided as a fixed visual prior and shared across all steps. Thus, \(x_t'\) depends only on \(x_t\), which depends only on \(x_{t-1}\), and \(x_{t-1}'\) is derived directly from \(x_{t-1}\), encapsulating all relevant information from \(x_{t-1}\). Since \(x_t'\) depends on \(x_{t-1}'\) only through the intermediate state \(x_t\), any dependency on earlier states (\(x_{t-2}', x_{t-2}, \ldots\)) is mediated by \(x_{t-1}'\). Each Gaussian noise term (\(\epsilon_{t-1}^*\), \(\epsilon_{t-1}\), \(\epsilon_t^*\)) is independently sampled. This ensures there are no hidden dependencies between \(x_t'\) and earlier states beyond \(x_{t-1}'\).

Given the encapsulation of information and the independence of noise, the conditional dependency simplifies to:
\[
P(x_t' | x_0, x_1', \ldots, x_{t-1}') = P(x_t' | x_{t-1}').
\]
Thus, the sequence \(X = \{x_0, x_1', x_2', \ldots, x_t'\}\) satisfies the Markov property.

\subsection*{New sampling distribution}

$x_t'$ and $x_{t-1}'$ can be expressed as follows:

\begin{align}
x_t' &= \sqrt{\bar{\alpha}_t} x_0 + \sqrt{1-\bar{\alpha}_t} \, \sqrt{\eta} x_s + \sqrt{1-\bar{\alpha}_t} \, \bar{\epsilon}_t - \sqrt{1-\bar{\alpha}_t} \, \sqrt{\eta} \, \epsilon_t^* \\
x_{t-1}' &= \sqrt{\bar{\alpha}_{t-1}} x_0 + \sqrt{1-\bar{\alpha}_{t-1}} \, \sqrt{\eta} x_s + \sqrt{1-\bar{\alpha}_{t-1}} \, \bar{\epsilon}_{t-1} - \sqrt{1-\bar{\alpha}_{t-1}} \, \sqrt{\eta} \, \epsilon_{t-1}^*  \\
x_t' &= x_t + \sqrt{1-\bar{\alpha}_t} \, \sqrt{\eta} x_s - \sqrt{1-\bar{\alpha}_t} \, \sqrt{\eta} \cdot \epsilon_t^* \\
    &= \sqrt{\alpha_t} x_{t-1} + \sqrt{1-\bar{\alpha}_t} \, \sqrt{\eta} x_s + \sqrt{1-\alpha_t} \, \epsilon_{t-1} - \sqrt{1-\bar{\alpha}_t} \, \sqrt{\eta} \epsilon_t^* \\
    &= \sqrt{\alpha_t} \left( x_{t-1}' - \sqrt{1-\bar{\alpha}_{t-1}} \, \sqrt{\eta} x_s + \sqrt{1-\bar{\alpha}_{t-1}} \, \sqrt{\eta} \epsilon_{t-1}^* \right) \\ \nonumber
    & \quad + \sqrt{1-\alpha_t} \, \epsilon_{t-1} - \sqrt{1-\bar{\alpha}_t} \, \sqrt{\eta} \epsilon_t^* + \sqrt{1-\bar{\alpha}_t} \, \sqrt{\eta} x_s \\
    &= \sqrt{\alpha_t} x_{t-1}' + \left( \sqrt{1-\bar{\alpha}_t} - \sqrt{\alpha_t} \sqrt{1-\bar{\alpha}_{t-1}} \right) \sqrt{\eta} x_s  \\ \nonumber
    & \quad + \sqrt{\alpha_t} \, \sqrt{1-\bar{\alpha}_{t-1}} \, \sqrt{\eta} \epsilon_{t-1}^* + \sqrt{1-\alpha_t} \, \epsilon_{t-1} - \sqrt{1-\bar{\alpha}_t} \, \sqrt{\eta} \epsilon_t^* .
\end{align}
Therefore,
\begin{align}
q(x_t' | x_0) &= \mathcal{N}\left(x_t'; \sqrt{\bar{\alpha}_t} x_0 + \sqrt{1-\bar{\alpha}_t} \, \sqrt{\eta} x_s, \, \left(1+\eta\right)(1-\bar{\alpha}_t) \mathbf{I}\right)  \\
q(x_{t-1}' | x_0) &= \mathcal{N}\left(x_{t-1}'; \sqrt{\bar{\alpha}_{t-1}} x_0 + \sqrt{1-\bar{\alpha}_{t-1}} \, \sqrt{\eta} x_s, \, (1+\eta)(1-\bar{\alpha}_{t-1}) \mathbf{I}\right)  \\
q(x_t' | x_{t-1}') &= \mathcal{N}\left(x_t'; \, \sqrt{\alpha_t} x_{t-1}' + \left(\sqrt{1-\bar{\alpha}_t} - \sqrt{\alpha_t} \sqrt{1-\bar{\alpha}_{t-1}} \right) \sqrt{\eta} x_s, \, \left((1+\alpha_t-2\bar{\alpha}_t) \eta + 1-\alpha_t \right) \mathbf{I}\right)  
\end{align}
Bayes' theorem can calculate the posterior distribution:
\begin{center}
\resizebox{\linewidth}{!}{ % 缩放公式
$
\begin{aligned}
q(x_{t-1}' | x_t', x_0) &= \frac{q(x_t' | x_{t-1}', x_0) \, q(x_{t-1}' | x_0)}{q(x_t' | x_0)} \\
&\propto \exp \left\{
    -\frac{1}{2} \left[
        \frac{\left(x_t' - \sqrt{\alpha_t} x_{t-1}’ - \left(\sqrt{1-\bar{\alpha}_t} - \sqrt{\alpha_t}\sqrt{1-\bar{\alpha}_{t-1}}\right)\sqrt{\eta} x_s\right)^2}{(1+\alpha_t-2\bar{\alpha}_t) \eta + 1-\alpha_t} \right. + \frac{\left(x_{t-1}' - \sqrt{\bar{\alpha}_{t-1}} x_0 - \sqrt{1-\bar{\alpha}_{t-1}} \sqrt{\eta} x_s\right)^2}{(1+\eta)(1-\bar{\alpha}_{t-1})}  \left. - \frac{\left(x_t' - \sqrt{\bar{\alpha}_t} x_0 - \sqrt{1-\bar{\alpha}_t} \sqrt{\eta} x_s\right)^2}{(1+\eta)(1-\alpha_t)} 
    \right] \right\} &\\
&= \exp \left\{ -\frac{1}{2} \left[
        \frac{\alpha_t x_{t-1}^2 - 2 \sqrt{\alpha_t} x_t' x_{t-1}' + 2 \sqrt{\alpha_t} \left( \sqrt{1-\bar{\alpha}_t} - \sqrt{\alpha_t}\sqrt{1-\bar{\alpha}_{t-1}} \right) \sqrt{\eta} x_s x_{t-1}'}{(1+\alpha_t-2\bar{\alpha}_t) \eta + 1-\alpha_t} + \frac{{x_{t-1}'}^2 - 2 \sqrt{\bar{\alpha}_{t-1}} x_0 x_{t-1}' - 2 \sqrt{1-\bar{\alpha}_{t-1}}\sqrt{\eta} x_s x_{t-1}'}{(1+\eta)(1-\bar{\alpha}_{t-1})} + \mathbf{C}(x_t',x_0)
    \right] \right\}   &\\
&\propto \exp \left\{ -\frac{1}{2} \left[
    \frac{\alpha_t}{(1+\alpha_t-2\bar{\alpha}_t) \eta + 1-\alpha_t}{x_{t-1}'}^2
    - 2 \frac{\sqrt{\alpha_t} x_t' - \sqrt{\alpha_t} \left( \sqrt{1-\bar{\alpha}_t} - \sqrt{\alpha_t}\sqrt{1-\bar{\alpha}_{t-1}} \right) \sqrt{\eta} x_s}{(1+\alpha_t-2\bar{\alpha}_t) \eta + 1-\alpha_t} x_{t-1}' + \frac{1}{(1+\eta)(1-\bar{\alpha}_{t-1})}{x_{t-1}'}^2
    -2 \frac{ \sqrt{\bar{\alpha}_{t-1}} x_0 + \sqrt{1-\bar{\alpha}_{t-1}} \sqrt{\eta} x_s}{(1+\eta)(1-\bar{\alpha}_{t-1})} x_{t-1}'
    \right] \right\}  &\\
&= \exp \left\{ -\frac{1}{2} \left[
    \frac{1-\bar{\alpha}_t+(1+2\alpha_t -3\bar{\alpha}_t)\eta}{\left((1+\alpha_t-2\bar{\alpha}_t) \eta + 1-\alpha_t\right)(1+\eta)(1-\bar{\alpha}_{t-1})}{x_{t-1}'}^2  -2 \left[
    \frac{\sqrt{\alpha_t} x_t' - \sqrt{\alpha_t} \left( \sqrt{1-\bar{\alpha}_t} - \sqrt{\alpha_t}\sqrt{1-\bar{\alpha}_{t-1}} \right) \sqrt{\eta} x_s}{(1+\alpha_t-2\bar{\alpha}_t) \eta + 1-\alpha_t} + \frac{\sqrt{\bar{\alpha}_{t-1}} x_0 + \sqrt{1-\bar{\alpha}_{t-1}} \sqrt{\eta} x_s}{(1+\eta)(1-\bar{\alpha}_{t-1})}\right]x_{t-1}' \right]\right\}  &\\
&= \exp \left\{ -\frac{1}{2} \frac{1}{\frac{\left((1+\alpha_t-2\bar{\alpha}_t) \eta + 1-\alpha_t\right)(1+\eta)(1-\bar{\alpha}_{t-1})}{1-\bar{\alpha}_t+(1+2\alpha_t -3\bar{\alpha}_t)\eta}} \left[
    {x_{t-1}'}^2  -2\frac{\left[\quad\cdots \quad\right]{\left((1+\alpha_t-2\bar{\alpha}_t) \eta + 1-\alpha_t\right)(1+\eta)(1-\bar{\alpha}_{t-1})}}{1-\bar{\alpha}_t+(1+2\alpha_t -3\bar{\alpha}_t)\eta} x_{t-1}'  \right] \right\}  &\\
&= \exp \left\{ -\frac{1}{2} \frac{1}{\frac{\left((1+\alpha_t-2\bar{\alpha}_t) \eta + 1-\alpha_t\right)(1+\eta)(1-\bar{\alpha}_{t-1})}{1-\bar{\alpha}_t+(1+2\alpha_t -3\bar{\alpha}_t)\eta}} \left[
    {x_{t-1}'}^2 -2\frac{\sqrt{\alpha_t}(1-\bar{\alpha}_{t-1})(1+\eta)x_t' + \sqrt{\bar{\alpha}_{t-1}} \left((1-\alpha_t-2\bar{\alpha}_t)\eta + 1-\alpha_t \right)x_0 + f(\eta)}{1-\bar{\alpha}_t+(1+2\alpha_t -3\bar{\alpha}_t)\eta} x_{t-1}' \right] \right\} &\\
&\propto \mathcal{N}\left(x_{t-1}'; \, \underbrace{\frac{\sqrt{\alpha_t}(1-\bar{\alpha}_{t-1})(1+\eta)x_t'  + \sqrt{\bar{\alpha}_{t-1}} \left((1-\alpha_t-2\bar{\alpha}_t)\eta + 1-\alpha_t \right)x_0 + f(\eta)}{1-\bar{\alpha}_t+(1+2\alpha_t -3\bar{\alpha}_t)\eta}}_{\mu_q(x_t', x_0)}, \, \underbrace{\frac{\left((1+\alpha_t-2\bar{\alpha}_t) \eta + 1-\alpha_t\right)(1+\eta)(1-\bar{\alpha}_{t-1})}{1-\bar{\alpha}_t+(1+2\alpha_t -3\bar{\alpha}_t)\eta} \mathbf{I}}_{\Sigma_q(t)}\right)
\end{aligned}
$
}
\end{center}

The function $f(\eta)$ represents the independent term dependent on $\eta$, which can be expressed as:
\begin{equation}
    f(\eta) = \sum_{k=1}^3 A_k \sqrt{\eta}^k,
\end{equation}
where $A_k$ are coefficients, determined by variables such as $\alpha_t$, $\alpha_{t-1}$, $\bar{\alpha}_t$, $\bar{\alpha}_{t-1}$, and $x_s$.

\subsection*{Model parametrization}
The mean of $q(x_{t-1}' | x_t', x_0)$ is:
\begin{equation}
    \mu_q(x_t', x_0) = \frac{\sqrt{\alpha_t}(1-\bar{\alpha}_{t-1})(1+\eta)x_t' + \sqrt{\bar{\alpha}_{t-1}} \left((1-\alpha_t-2\bar{\alpha}_t)\eta + 1-\alpha_t \right)x_0 + f(\eta)}{1-\bar{\alpha}_t+(1+2\alpha_t -3\bar{\alpha}_t)\eta}
\end{equation}
From the derivation of the forward process $q(x_t' | x_0)$, we have an equivalent interpretation of $x_0$ expressed as:
\begin{equation}
    x_0 = \frac{x_t - \sqrt{1 - \bar{\alpha}_t} \cdot \sqrt{1 - \eta} \cdot \epsilon - \sqrt{1 - \bar{\alpha}_t} \cdot \sqrt{\eta} \cdot x_s}{\sqrt{\bar{\alpha}_t}}
\end{equation}

Let
\begin{equation}
    \tau = \sqrt{1+\eta}\epsilon + \sqrt{\eta}x_s,
\end{equation}
the alternate parameterization for true denoising transition mean $\mu_q(x_t', x_0)$ is:
\begin{align}
    \mu_q(x_t', t) &= \frac{\sqrt{\alpha_t}(1-\bar{\alpha}_{t-1})(1+\eta)x_t' + \sqrt{\bar{\alpha}_{t-1}} \left((1-\alpha_t-2\bar{\alpha}_t)\eta + 1-\alpha_t \right) \frac{x_t' - \sqrt{1-\bar{\alpha}_t}\tau}{\sqrt{\bar{\alpha}_t}} + f(\eta)}{1-\bar{\alpha}_t+(1+2\alpha_t -3\bar{\alpha}_t)\eta}  &\\
    &= \frac{
    \sqrt{\bar{\alpha}_{t-1}}(1 - \bar{\alpha}_t)x'_t 
    + (C_1)x'_t 
    - \sqrt{\bar{\alpha}_{t-1}}\sqrt{1 - \bar{\alpha}_t}(1 - \alpha_t)\tau 
    + (C_2)\tau + f(\eta)
}{
    \sqrt{\bar{\alpha}_t}\left(1-\bar{\alpha}_t+(1+2\alpha_t -3\bar{\alpha}_t)\eta \right)
},
\end{align}
where $C_1$ and $C_2$ are also coefficients, with each term multiplied with $\eta$.

We parameterize $\hat{\tau}_\theta(x_t',t)$ by neural networks that seek to predict both \emph{underlying noise} and the \emph{prior’s effect} from noisy image $x_t'$. Then, the optimization problem simplifies to:

\begin{equation}
    \mathbf{L}_\tau(\theta) = \mathbb{E}\left[||\tau - \hat{\tau}_\theta(x_t',t)||^2\right].
\end{equation}
During inference, let $\eta=0$, then:
\begin{align}
    \mu_q(x_t', t) &= \mu_q(x_t, t) \text{\quad in DDPM},  &\\
    \Sigma_q(t) &= \Sigma_q(t) \text{\quad in DDPM} 
\end{align}
We can still utilize the DDPM sampling formula to generate an image $\hat{x}$ from Gaussian noise without requiring an additional visual prior $x_s$.

\noindent\textbf{The proof of $\mu_q(x_t, t)$ in DDPM.}
\begin{align}
    \mu_\theta(x_t, t) &= \frac{\sqrt{\alpha_t} (1 - \bar{\alpha}_{t-1}) x_t + \sqrt{\bar{\alpha}_{t-1}} (1 - \alpha_t) \frac{x_t - \sqrt{1 - \bar{\alpha}_t} \epsilon}{\sqrt{\bar{\alpha}_t}}}{1 - \alpha_t}  &\\
    &= \frac{\sqrt{\alpha_t}\sqrt{\bar{\alpha}_t} (1 - \bar{\alpha}_{t-1}) x_t + \sqrt{\bar{\alpha}_{t-1}} (1 - \alpha_t) (x_t - \sqrt{1-\bar{\alpha}_t}\epsilon) }{{\sqrt{\bar{\alpha}_t}}(1 - \alpha_t)}  &\\
    &= \frac{\alpha_t\sqrt{\bar{\alpha}_{t-1}} (1 - \bar{\alpha}_{t-1}) x_t + \sqrt{\bar{\alpha}_{t-1}} (1 - \alpha_t)x_t - \sqrt{\bar{\alpha}_{t-1}}\sqrt{1-\bar{\alpha}_t} (1 - \alpha_t)\epsilon}{{\sqrt{\bar{\alpha}_t}}(1 - \alpha_t)}  &\\
    &= \frac{\sqrt{\bar{\alpha}_{t-1}}(1-\bar{\alpha}_t) - \sqrt{\bar{\alpha}_{t-1}}\sqrt{1-\bar{\alpha}_t} (1 - \alpha_t)\epsilon}{{\sqrt{\bar{\alpha}_t}}(1 - \alpha_t)}  &\\
    &= \frac{1}{\sqrt{\alpha_t}} \left(x_t - \frac{1 - \alpha_t}{\sqrt{1 - \bar{\alpha}_t}} \epsilon\right)
\end{align}
$ \mu_q(x_t', t)$ equals equation (31) when $\eta=0$.

\section*{Stroke parameter fitting (Relevant to Section \ref{sec:control} in the Main Text)}

In differentiable rasterizer \cite{li2020differentiable}, we set the fitting conditions to [path=1, segment=1], ensuring that one cubic Bézier curve corresponds to a single stroke image. Due to the random initialization of the curves, gradient vanishing often occurs during updates when the curve has no overlap with the stroke. To address this, we initially added an optimal transport loss $Loss_{OT}$ for minimizing the distance between distributions, as proposed by \cite{zou2021stylized}, to the pixel loss optimization function. However, this approach proved ineffective. In fact, we also tested $Loss_{OT}$ in \cite{zou2021stylized} and observed that when the initial sampling points of a stroke did not overlap with the object in the image (using white-background images), the strokes failed to move toward the object and instead gradually disappeared during backpropagation.

\begin{figure*}[t]
\centering
\begin{minipage}[t]{0.68\textwidth}
    \centering
    \resizebox{\textwidth}{!}{%
    \begin{tabular}{lllllllllcllll}
    \toprule
    Dimension & $p_{0,x}$ & $p_{0,y}$ & $p_{1,x}$ & $p_{1,y}$ & $p_{2,x}$ & $p_{2,y}$ & $p_{3,x}$ & $p_{3,y}$ 
    & R & G & B & Opacity & Width \\
    \midrule
    Min. & 12 & 22 & -100 & -195 & -84 & -140 & -9 & 22 & 0 & 0 & 0 & 0 & 6 \\
    Max. & 268 & 273 & 450 & 399 & 465 & 448 & 267 & 305 & 255 & 255 & 255 & 1 & 106 \\
    \bottomrule
    \end{tabular}
    }
    \captionof{table}{The value ranges of stroke parameters for each dimension at an image resolution of $295 \times 295$.}
    \label{tab:param}
\end{minipage}
\hfill
\begin{minipage}[t]{0.3\textwidth}
    \centering
    \includegraphics[width=\linewidth]{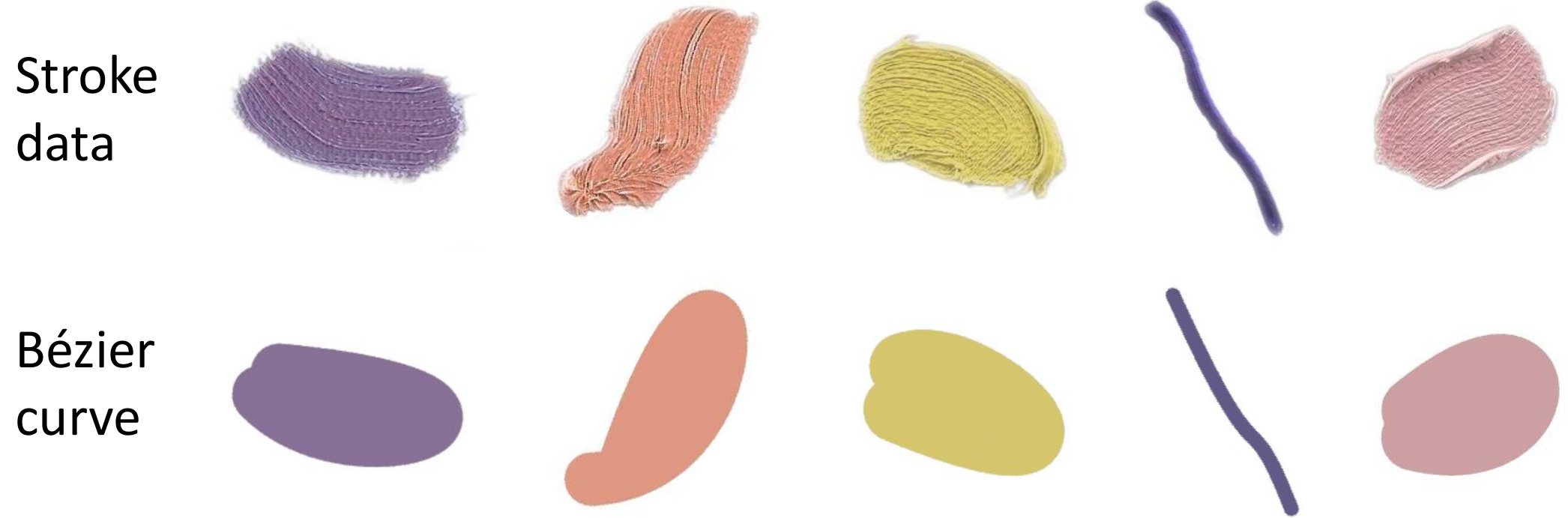}
    \captionof{figure}{Bézier curves fitted in the same shape as our stroke data.}
    \label{fig:svg}
\end{minipage}
\end{figure*}

Ultimately, we adopted a more practical solution: detecting the regions with strokes in the image and initializing the curve in those regions. Each stroke's parameters were optimized over 300 iterations to complete the parameterization. Some fitted examples are shown in Figure \ref{fig:svg}. We present the range of values for each parameter dimension in Table \ref{tab:param}, based on images with a resolution of 295$\times$295 pixels.

\section*{Code for Ranking Loss (Relevant to Section \ref{sec:pipeline} in the Main Text)}

Listing \ref{alg:code} presents the core code for calculating the ranking loss of a single set of strokes on a canvas patch. Batch operations are omitted for simplicity.

\begin{Code}[t]
\label{alg:code}
\centering
\footnotesize
\begin{lstlisting}[
  caption={A simplified code of ranking loss calculation for one stroke sequence.}, 
  label={alg:code}, 
  frame=lines,
  aboveskip=0.5em, 
  belowskip=0.5em
]
from scipy.special import comb
def cal_ranking_loss(pred_rank, gt_order, margin=0.125):
    dif_gt = gt_order - gt_order.t()
    dif_pred = pred_rank - pred_rank.t()
    # All pairwise combinations: positions with value 1 indicate that the row corresponds
    # to a ground truth rendering order earlier than the column.
    mask = (dif_gt < 0).float()    
    loss_matric = torch.max(torch.tensor(0.0), (dif_pred - dif_gt * margin) * mask)
    loss = loss_matric.sum() / comb(pred_rank.shape[0], 2)
    return loss

loss_r = 0.0
param, _ = stroke_predictor(canvas1, canvas2)
# matching_idx comes from Hungarian matching
# -1 takes the ranking dimension
pred_rank = param[matching_idx, -1]
gt_order = gt_param[valid, -1]
loss_r = cal_ranking_loss(pred_rank, gt_order)
\end{lstlisting}
\end{Code}

\begin{figure*}
    \centering
    \includegraphics[width=0.95\linewidth]{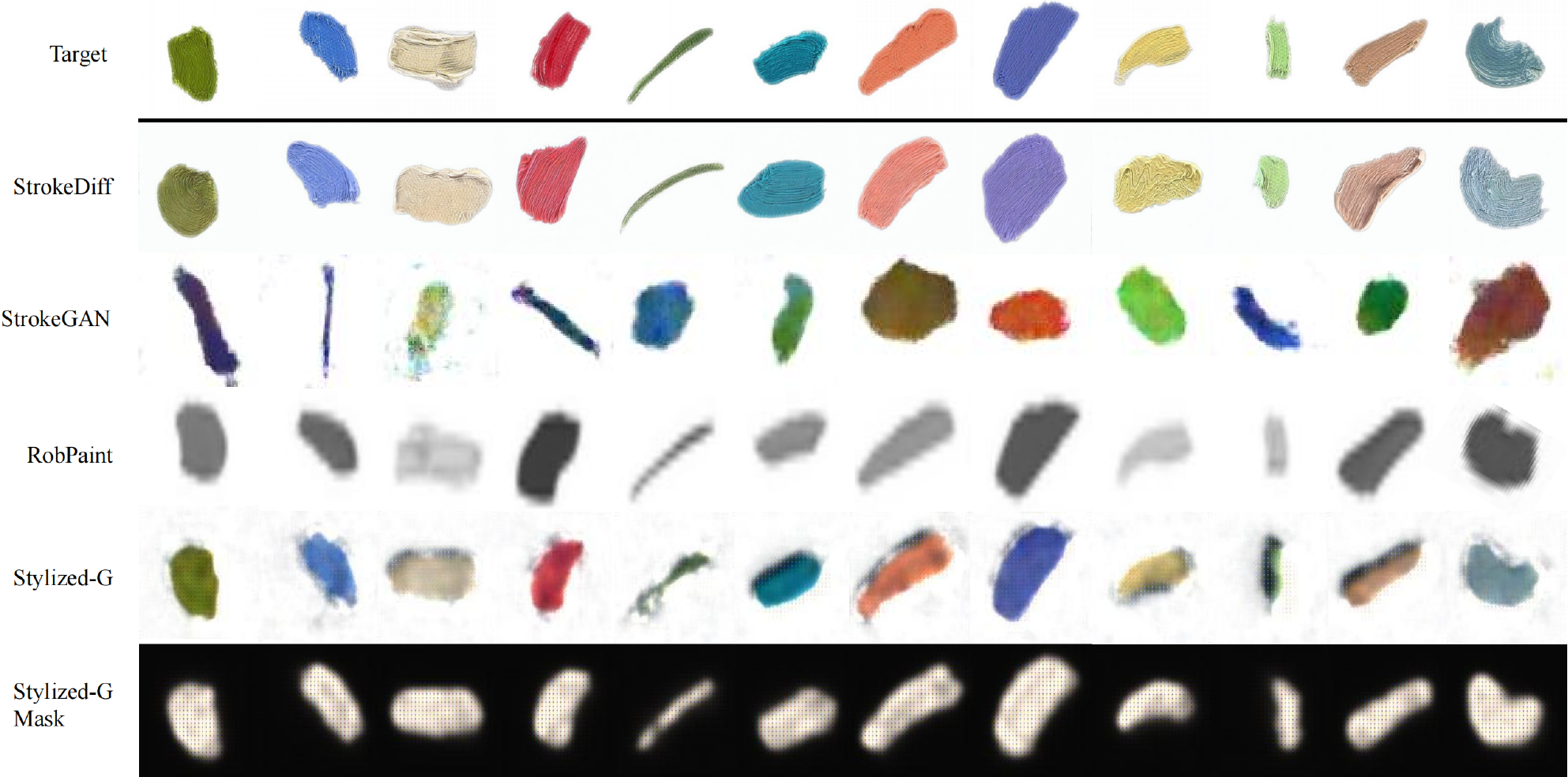}
    \caption{Comparisons between our stroke generator and StrokeGAN \cite{wang2023stroke}, Stylized \cite{zou2021stylized} and RobPaint \cite{bidgoli2020artistic}.}
    \label{fig:compar}
\end{figure*}

\section*{Stroke generation comparison (Relevant to Section \ref{sec:results_comp} in the Main Text)}
\label{supp:compare}

In Figure \ref{fig:compar}, we present a qualitative comparison of our results with those generated by other stroke generators. For StrokeGAN, we trained the model for 1000 epochs and selected the best-performing weights from the 700th epoch for comparison. For Stylized-G, we followed the original setup of the method, where stroke parameters excluding RGB were input to the "shading net" to predict the mask, and all parameters were fed into the "rasterization net" to predict the foreground. The model was trained for 370 epochs. For RobPaint, we used the default setting with an output size of $1\times16\times32$ and trained for 255 epochs.

\section*{User study (Relevant to Section \ref{sec:user_study} in the Main Text)}

Painting quality was evaluated using four questions:
Style (“To what extent does the artwork emphasize its oil painting style?”),
Aesthetics (“As an artwork, how valuable are they from an aesthetic point of view?”),
Texture (“How well are the oil-painting brushstroke textures expressed in the artwork?”),
and Content (“From the photograph to the painting, how well is the content applied in the artwork?”).
Participants evaluated paintings produced by six methods, with five examples per method.

\end{document}